\documentclass[journal]{IEEEtran}
 \usepackage{graphicx}

 \graphicspath{{fig/}}

\newif\ifReview

\newif\ifShortVersion

\newif\ifWithContribution
\ifReview
\WithContributionfalse
\else
\WithContributiontrue
\fi

\newif\ifNotFullExperiment
\NotFullExperimenttrue

\newif\ifDummy

\newif\ifShowTheoremLTwo

\newif\ifGraphWithoutSign

\newif\ifShowTheoremMIPS
\ShowTheoremMIPStrue

\newif\ifWithSectionNumber
\WithSectionNumbertrue

\newif\ifOmitAdditionalTheoreticalResult
\OmitAdditionalTheoreticalResulttrue

\newif\ifOmitAdditionalTexMexResult
\OmitAdditionalTexMexResulttrue

\newif\ifOmitAdditionalCFResult
\OmitAdditionalCFResulttrue

\newif\ifOmitAdditionalIRResult
\OmitAdditionalIRResulttrue

\newif\ifOmitCovertreeAlgorithm
\OmitCovertreeAlgorithmtrue

\newif\ifOmitBiography
\OmitBiographytrue

\newif\ifOmitAppendix

\newif\ifSuppressMemo

\ifSuppressMemo
\newcommand{\memo}[1]{}
\else
\usepackage{color}
\newcommand{\memo}[1]{{\bf \color{red}{#1}} \color{black}}
\fi

\usepackage[utf8]{inputenc} 
\usepackage[T1]{fontenc}    
\usepackage{hyperref}       
\usepackage{url}            
\usepackage{booktabs}       
\usepackage{amsfonts}       
\usepackage{nicefrac}       
\usepackage{microtype}      

 \usepackage[update]{epstopdf}
\usepackage{subfigure} 
\usepackage{amssymb}
\usepackage{amsmath}
\usepackage{amsbsy}
\usepackage{latexsym}
\usepackage{multirow}

\usepackage{algorithm}
\usepackage{algpseudocode,algorithmicx}

\usepackage{nkj,nkj_e}

\newcommand{\citep}{\cite}
\newcommand{\citet}{\cite}

\usepackage{lipsum}

\newcommand{\ctlevel}{i}
\newcommand{\pnorm}[2]{\|#1\|_{\raisebox{-2pt}{\tiny\ensuremath{#2}}}}
\newcommand{\pnormAdjustedBrackets}[2]{\left\|#1\right\|_{\raisebox{-2pt}{\tiny\ensuremath{#2}}}}
\newcommand{\storeCat}{\ensuremath{\text{stor}_{\mathrm{CAT}}}}
\newcommand{\CATError}[1]{\mathcal{E}_{#1}}
\usepackage{xspace}
\newcommand{\LTwoDissimilarity}{squared-L2-distance\xspace}
\allowdisplaybreaks

\ifCLASSINFOpdf
\else
\fi
\begin{document}
%
\title{Sharing Hash Codes for Multiple Purposes}
%
%
%

\author{
Wiktor Pronobis$^{\dagger}$, Danny Panknin$^{\dagger}$, Johannes Kirschnick$^{\dagger}$, Vignesh Srinivasan, Wojciech Samek, \\
Volker Markl, Manohar Kaul, Klaus-Robert M{\"u}ller$^{*}$, and Shinichi Nakajima

\ifWithContribution
\thanks{
$\dagger$ contributed equally.
}
\fi
\thanks{$^*$corresponding author (email: Klaus-robert.mueller@tu-berlin.de)}%
\thanks{Wikor Pronobis, Danny Panknin, Klaus-Robert M{\"u}ller, and Shinichi Nakajima are with Technische Unversit{\"a}t Berlin, Machine Learning Group, Marchstr. 23, 10587 Berlin, Germany.}
\thanks{Johannes Kirschnick and Volker Markl are with DFKI, Language Technology Lab, Alt-Moabit 91c, Berlin, Germany.}%
\thanks{Vignesh Srinivasan and Wojciech Samek are with Fraunhofer HHI.}
\thanks{Volker Markl is with Technische Unversit{\"a}t Berlin, Database Systems and Information Management Group, Einsteinufer 17, 10587 Berlin, Germany.}
\thanks{Manohar Kaul is with IIT Hyderabad.}
\thanks{Klaus-Robert M{\"u}ller is with Korea University, and with Max Planck Society.}%
\thanks{Wojciech Samek, Volker Markl, Klaus-Robert M{\"u}ller, and Shinichi Nakajima are with Berlin Big Data Center, 10587 Berlin, Germany.}
}

%
%

\markboth{Journal of \LaTeX\ Class Files,~Vol.~14, No.~8, August~2015}%
{Pronobis \MakeLowercase{\textit{et al.}}: Multiple Purpose Locality Sensitive Hashing}
%



\maketitle

\begin{abstract}
Locality sensitive hashing (LSH) is a powerful tool for sublinear-time approximate nearest neighbor search,
and a variety of hashing schemes have been proposed for different dissimilarity measures.
However, hash codes significantly depend on the dissimilarity,
which prohibits users from adjusting the dissimilarity at query time. 
In this paper, 
we propose \emph{multiple purpose LSH} (mp-LSH) which shares the hash codes
for different dissimilarities.
mp-LSH supports L2, cosine, and inner product dissimilarities,
and their corresponding weighted sums,
where the weights can be adjusted \emph{at query time}.
It also allows us to modify the importance of pre-defined groups of features.
Thus, mp-LSH enables us, for example, to retrieve similar items to a query with the user preference taken into account,
to find a similar material to a query with some properties (stability, utility, etc.) optimized,
and to turn on or off a part of multi-modal information (brightness, color, audio, text, etc.) in image/video retrieval.
We theoretically and empirically analyze the performance of three variants of mp-LSH,
and demonstrate their usefulness on real-world data sets.
\end{abstract}


%
\IEEEpeerreviewmaketitle

\section{Introduction}

Large amounts of data are being collected every day in the sciences and industry. Analysing such truly big data sets even by linear methods can become infeasible, thus sublinear methods such as locality sensitive hashing (LSH) have become an important analysis tool. For some data collections, the purpose can be clearly expressed from the start, for example, text/image/video/speech analysis or recommender systems.
In other cases such as drug discovery or the material genome project, 
the ultimate query structure to such data may still not be fully fixed. In other words, measurements, simulations or observations may be recorded without being able to spell out the full specific purpose (although the general goal: better drugs, more potent materials is clear). Motivated by the latter case, we consider how one can use LSH schemes without defining any specific dissimilarity at the data acquisition and pre-processing phase.

LSH, one of the key technologies for big data analysis, enables approximate nearest neighbor search (ANNS) in \emph{sublinear} time \citep{Indyk98,Wang14}.
With LSH functions for a required dissimilarity measure in hand, each data sample is assigned to a \emph{hash bucket} in the pre-prosessing stage.
At runtime, ANNS with theoretical guarantees can be performed by restricting the search to the samples that lie within the hash bucket, to which the query point is assigned, along with the samples lying in the neighbouring buckets.

 A challenge in developing LSH without defining specific purpose is that
the existing LSH schemes, designed for different dissimilarity measures, provide significantly different hash codes.
 Therefore, a naive realization requires us to prepare the same number of hash tables as the number of possible target dissimilarities,
 which is not realistic if we need to adjust the importance of multiple criteria.
\ifShortVersion
\else
In this paper, we propose three variants of multiple purpose LSH (mp-LSH),
which support L2, cosine, and inner product (IP) dissimilarities,
and their weighted sums, where the weights can be adjusted at query time.

The first proposed method, called mp-LSH with vector augmentation (mp-LSH-VA),
maps the data space into an augmented vector space,
so that the \LTwoDissimilarity in the augmented space matches the required dissimilarity measure up to a constant.
This scheme can be seen as an extension of 
recent developments of LSH for maximum IP search (MIPS) \citep{Shrivastava14,Bachrach14,Shrivastava15,Neyshabur15}.
The significant difference from the previous methods is that
our method is designed to modify the dissimilarity by changing the augmented query vector.
We show that mp-LSH-VA is locality sensitive for L2 and IP dissimilarities and their weighted sums.
However, its performance for the L2 dissimilarity is significantly inferior to the standard L2-LSH \citep{Datar04}.
In addition, mp-LSH-VA does not support the cosine-distance.

Our second proposed method, called mp-LSH with code concatenation (mp-LSH-CC),
concatenates the hash codes for L2, cosine, and IP dissimilarities,
and constructs a special structure, called \emph{cover tree} \citep{Bustos12},
which enables efficient NNS with the weights for the dissimilarity measures controlled by adjusting the metric in the code space.
Although mp-LSH-CC is conceptually simple and its performance is guaranteed by the original LSH scheme for each dissimilarity,
it is not memory efficient, which also results in increased query time.

Considering the drawbacks of the aforementioned two variants
led us to our final and recommended proposal, called mp-LSH with code augmentation and transformation (mp-LSH-CAT).
It supports L2, cosine, and IP dissimilarities
by augmenting the hash codes, instead of the original vector.
mp-LSH-CAT is memory efficient, since it shares most information over the hash codes for different dissimilarities,
so that the augmentation is minimized.


We theoretically and empirically analyze the performance of mp-LSH methods,
and demonstrate their usefulness on 
real-world data sets.
Our mp-LSH methods also allow us to modify the importance of pre-defined groups of features.
Adjustability of the dissimilarity measure at query time is not only useful in the absence of future analysis plans, but also
 applicable to multi-criteria searches. The following lists some sample applications of multi-criteria queries in diverse areas:
\begin{enumerate}

\item In recommender systems, suggesting items which are similar to a user-provided query and also match the user’s preference.

\item In material science, finding materials which are similar to a query material
and also possess desired properties
such as stability, conductivity, and medical utility. 

\item In video retrieval, we can adjust the importance of multimodal information such as brightness, color, audio, and text
at query time.

\end{enumerate}

\fi


\section{Background}

In this section, we briefly overview previous locality sensitive hashing (LSH) techniques.
%

Assume that we have a sample pool $\mcX = \{\bfx^{(n)} \in  \mathbb{R}^{L}\}_{n=1}^{N}$
in $L$-dimensional space.
Given a query $\bfq \in \mathbb{R}^{L} $,
nearest neighbor search (NNS) solves the 
\ifShortVersion
\else
following problem:
\begin{align}
\bfx^* =   \argmin_{\bfx \in \mcX} \mathcal{L}(\bfq, \bfx),
\label{eq:NNSProblem}
\end{align}
\fi
where $ \mathcal{L}(\cdot, \cdot)$ is a dissimilarity measure.
A naive approach computes the dissimilarity from the query to all samples, and then chooses the most similar samples,
which takes $O(N)$ time.
On the other hand, approximate NNS can be performed in sublinear time.
We define the following three terms:
\begin{definition} 
\label{def:NN}
($S_0$-near neighbor)
For $S_0 > 0$,
$\bfx$ is called $S_0$-near neighbor of $\bfq$, if $\mathcal{L}(\bfq, \bfx) \leq S_0$.
\end{definition}
\begin{definition} 
\label{def:CANNS}
($c$-approximate nearest neighbor search)
Given $S_0 >0$, $\delta > 0$, and $c > 1$, 
$c$-approximate nearest neighbor search ($c$-ANNS) reports some $cS_0$-near neighbor of $\bfq$ with probability $1 - \delta$,
if there exists an $S_0$-near neighbor of $\bfq$ in $\mcX$.
\end{definition}
%
\begin{definition} 
\label{def:LSH}
(Locality sensitive hashing)
A family $\mcH = \{h: \mathbb{R}^{L} \to \mcK\}$ of functions is called $(S_0, c S_0, p_1, p_2)$-sensitive for a dissimilarity measure $\mathcal{L}: \mathbb{R}^{L} \times \mathbb{R}^{L} \to \mathbb{R}$,
if the following two conditions hold for any $\bfq, \bfx \in \mathbb{R}^{L}$:
\begin{align}
& \bullet \mbox{ if } \mathcal{L}(\bfq, \bfx) \leq S_0 \mbox{ then } \mathbb{P} \left(h(\bfq) = h(\bfx) \right) \geq p_1,
\notag\\
& \bullet \mbox{ if } \mathcal{L}(\bfq, \bfx) \geq c S_0 \mbox{ then } \mathbb{P} \left(h(\bfq) = h(\bfx) \right) \leq p_2,
\notag
\end{align}
where $\mathbb{P}(\cdot)$ denotes the probability of the event (with respect to the random draw of 
hash functions).
\end{definition}
Note that $p_1 > p_2$ is required for LSH to be useful.
The image $\mcK$ of hash functions is typically binary or integer.
The following proposition guarantees that 
locality sensitive hashing (LSH) functions enable $c$-ANNS in 
sublinear time.
\begin{proposition}
\label{prpt:CANNS}
\cite{Indyk98} 
Given a family of $(S_0, cS_0, p_1, p_2)$-sensitive hash functions, there exists an algorithm for $c$-ANNS
with $O(N^{\rho} \log N)$ query time and $O(N^{1 + \rho})$ space, where $\rho = \frac{\log p_1}{\log p_2} < 1$.
\end{proposition}


Below, we introduce three LSH families.
Let
$\mathcal{N}_L(\bfmu, \bfSigma)$ be the $L$-dimensional Gaussian distribution,
$\mathcal{U}_L(\alpha, \beta)$ be the $L$-dimensional uniform distribution with its support $[\alpha, \beta]$ for all dimensions,
and $\bfI_L$ be the $L$-dimensional identity matrix.
The sign function,
$\mathrm{sign}(\bfz): \mathbb{R}^H \mapsto \{-1, 1\}^H$,
 applies element-wise, giving $1$ for $z_h \geq 0$ and $-1$ for $z_h < 0$.
Likewise, the floor operator $\lfloor \cdot \rfloor$ applies element-wise for a vector.
We denote by $\sphericalangle(\cdot, \cdot)$ the angle between two vectors.
\begin{proposition} (L2-LSH)
\label{prpt:L2LSH}
\cite{Datar04} 
For the L2-distance
$\mcL_{\mathrm{L2}}(\bfq, \bfx) = \pnorm{\bfq - \bfx}{2}$,
the hash function
\Dummyfalse
\ifDummy
$h_{\bfa, b}^{\mathrm{L2}} (\bfx) 
= \textstyle
 \left\lfloor R^{-1}(\bfa^\T \bfx + b) \right\rfloor$,
\else
\begin{align}
h_{\bfa, b}^{\mathrm{L2}} (\bfx) 
&= \textstyle
 \left\lfloor R^{-1}(\bfa^\T \bfx + b) \right\rfloor,
\label{eq:L2LSH}
\end{align}
\fi
where $R > 0$ is a fixed real number, $\bfa \sim \mathcal{N}_L(\bfzero, \bfI_L)$,
and  $b \sim \mathcal{U}_1(0, R)$,
satisfies
$\mathbb{P} (h_{\bfa, b}^{\mathrm{L2}} (\bfq) = h_{\bfa, b}^{\mathrm{L2}} (\bfx)  )
= F_R^{\mathrm{L2}}(\mcL_{\mathrm{L2}}(\bfq, \bfx))$,
where
\Dummyfalse
\ifDummy
$F_R^{\mathrm{L2}}(d) = 1 - 2 \varPhi(-R / d) - \textstyle  \frac{2}{\sqrt{2 \pi} (R/d)} (1 - e^{-(R/d)^2/2} )$.
\else
\begin{align}
\textstyle 
F_R^{\mathrm{L2}}(d) &= 1 - 2 \varPhi(-R / d) - \textstyle  \frac{2}{\sqrt{2 \pi} (R/d)} \left(1 - e^{-(R/d)^2/2} \right).
\notag
\end{align}
\fi
Here, $\varPhi(z) = \int_{-\infty}^{z} \frac{1}{\sqrt{2 \pi}} e^{-\frac{y^2}{2}} dy$ is the 
standard cumulative Gaussian.
\end{proposition}

\begin{proposition} (sign-LSH)
\label{prpt:SignLSH}
\cite{Goemans95,Charikar02} 
For the cosine-distance $\mcL_{\mathrm{cos}}(\bfq, \bfx) = 1 - \cos \sphericalangle(\bfq, \bfx)= 1 -  \frac{ \bfq^\T  \bfx}{\| \bfq\|_2 \|\bfx\|_2}$,
the hash function
\Dummyfalse
\ifDummy
$h^{\mathrm{sign}}_{\bfa}(\bfx)
 = \mathrm{sign}(\bfa^\T \bfx)$,
\else
\begin{align}
h^{\mathrm{sign}}_{\bfa}(\bfx)
& = \mathrm{sign}(\bfa^\T \bfx),
\label{eq:SignLSH}
\end{align}
\fi
where 
$\bfa \sim \mathcal{N}_L(\bfzero, \bfI_L)$,
satisfies
$\mathbb{P} \left(h_{\bfa}^{\mathrm{sign}} (\bfq) = h_{\bfa}^{\mathrm{sign}} (\bfx)  \right)
= F^{\mathrm{sign}}(\mcL_{\mathrm{cos}}(\bfq, \bfx))$,
where
\Dummyfalse
\ifDummy
$F^{\mathrm{sign}}(d)  = \textstyle 1 - \frac{1}{\pi} \cos^{-1} (1-d)$.
\else
\begin{align}
F^{\mathrm{sign}}(d) & = \textstyle 1 - \frac{1}{\pi} \cos^{-1} (1-d).
\label{eq:FR.SignLSH}
\end{align}
\fi
\end{proposition}

\begin{proposition}  \citep{Neyshabur15} (simple-LSH)
\label{prpt:SimpleLSH}
Assume that 
the samples and the query are rescaled so that $\max_{\bfx \in \mcX} \|\bfx\|_2 \leq 1$, $\|\bfq\|_2 \leq 1$.
For the inner product dissimilarity 
$\mcL_{\mathrm{ip}}(\bfq, \bfx) = 1 - \bfq^\T  \bfx$
\ifShortVersion
\else
(with which the NNS problem \eqref{eq:NNSProblem} is called maximum IP search (MIPS)),
\fi
the asymmetric hash functions
\begin{align}
h^{\mathrm{smp-q}}_{\bfa}(\bfq)
&= h^{\mathrm{sign}}_{\bfa}(\widetilde{\bfq})
 = \mathrm{sign}(\bfa^\T \widetilde{\bfq}),
\label{eq:SimpleLSHQ}\\
& \qquad \qquad
\mbox{where}\qquad 
\widetilde{\bfq}
= \textstyle (\bfq; 0),
\notag\\
h^{\mathrm{smp-x}}_{\bfa}(\bfx)
&= h^{\mathrm{sign}}_{\bfa}(\widetilde{\bfx})
 = \mathrm{sign}(\bfa^\T \widetilde{\bfx}),
\label{eq:SimpleLSHX}\\
& \qquad \qquad
\mbox{where}\qquad 
\widetilde{\bfx}
= \textstyle (\bfx; \sqrt{1 - \|\bfx\|_2^2} ),
\notag
\end{align}
satisfy
$\mathbb{P} \left(h_{\bfa}^{\mathrm{smp-q}} (\bfq) = h_{\bfa}^{\mathrm{smp-x}} (\bfx)  \right)
= F^{\mathrm{sign}}(\mcL_{\mathrm{ip}}(\bfq, \bfx))$.
\end{proposition}

These three LSH methods above are standard and state-of-the-art
(among the data-independent LSH schemes) 
for each dissimilarity measure.
Although all methods involve the same random projection $\bfa^\T \bfx$,
the resulting hash codes are significantly different from each other.

\section{Proposed Methods and Theory}
\label{sec:ProposedMethod}

In this section, we first define the problem setting.
Then, we propose three LSH methods for multiple dissimilarity measures,
and conduct a theoretical analysis.

\subsection{Problem Setting}
\label{sec:ProblemSetting}

Similarly to the simple-LSH (Proposition~\ref{prpt:SimpleLSH}),
we rescale the samples so that $\max_{\bfx \in \mcX} \|\bfx\|_2 \leq 1$.
We also assume $\|\bfq\|_2 \leq 1$.%
\footnote{
This assumption is reasonable for L2-NNS if the size of the sample pool is sufficiently large,
and the query follows the same distribution as the samples.
For MIPS, the norm of the query can be arbitrarily modified,
and we set it to $\|\bfq\|_2 = 1$.
}
Let us assume multi-modal data,
where we can separate the feature vectors into $G$ groups, i.e.,
$\bfq = (\bfq_1; \ldots; \bfq_G)$, $\bfx = (\bfx_1; \ldots; \bfx_G)$.
For example, each group corresponds to monochrome, color, audio, and text features in video retrieval.
We also accept multiple queries $\{\bfq^{(w)}\}_{w=1}^W$ for a single retrieval task.
Our goal is to perform ANNS for the following dissimilarity measure,
which we call multiple purpose (MP) dissimilarity:
 \begin{align}
 &\mcL_{\mathrm{mp}}(\{\bfq^{(w)}\}, \bfx)
 = \textstyle \sum_{w=1}^W   \sum_{g=1}^G \bigg\{  \gamma_g^{(w)} \|\bfq_g^{(w)} - \bfx_g\|_2^2
 \notag \\
 & \hspace{-2mm}
 \textstyle
  + 2 \eta_g^{(w)} \left(1 -  \frac{\bfq_g^{(w) \T} \bfx_g}{\|\bfq_g^{(w)} \|_2 \|\bfx_g\|_2}\right)
 + 2 \lambda_g^{(w)}\left(1 - \bfq_g^{(w) \T} \bfx_g\right)\bigg\},
 \label{eq:Objective}
 \normalsize
 \end{align}
 where $\bfgamma^{(w)}, \bfeta^{(w)}, \bflambda^{(w)} \in \mathbb{R}_+^G$ are the feature weights such that $\sum_{w=1}^W \sum_{g=1}^G (\gamma_g^{(w)}+\eta_g^{(w)}+\lambda_g^{(w)}) = 1$.
In the single query case, where $W=1$,
setting $\bfgamma = (1/2, 0, 1/2, 0, \ldots, 0), \bfeta = \bflambda = (0, \ldots, 0)$ corresponds to L2-NNS based on the first and the third feature groups,
while setting
 $\bfgamma = \bfeta = (0, \ldots, 0), \bflambda = (1/2, 0, 1/2, 0, \ldots, 0)$ corresponds to MIPS on the same feature groups.
When we like to down-weight the importance of signal amplitude (e.g., brightness of image) of the $g$-th feature group,
we should increase the weight $\eta_g^{(w)}$ for the cosine-distance,
and decrease the weight $\gamma_g^{(w)}$ for the \LTwoDissimilarity.
 Multiple queries are useful when we mix NNS and MIPS,
 for which the queries lie in different spaces with the same dimensionality.
For example,
by setting $\bfgamma^{(1)}=  \bflambda^{(2)} = (1/4, 0, 1/4, 0, \ldots, 0), \bfgamma^{(2)} = \bfeta^{(1)}= \bfeta^{(2)} =\bflambda^{(1)}= (0, \ldots, 0)$,
we can retrieve items, which are close to the item query $\bfq^{(1)}$ and match the user preference query $\bfq^{(2)}$.
An important requirement for our proposal is that 
the weights $\{\bfgamma^{(w)}, \bfeta^{(w)}, \bflambda^{(w)}\}$ can be adjusted \emph{at query time}.

\ifShortVersion
\else
\subsection{Multiple purpose LSH with Vector Augmentation (mp-LSH-VA)}
\fi
\label{sec:mp-LSH-VA}

Our first method, called multiple purpose LSH with vector augmentation (mp-LSH-VA),
 is inspired by the research on asymmetric LSHs for MIPS \citep{Shrivastava14,Bachrach14,Shrivastava15,Neyshabur15},
 where the query and the samples are augmented with additional entries,
 so that the \LTwoDissimilarity in the augmented space coincides with the target dissimilarity up to a constant.
 A significant difference of our proposal from the previous methods 
 is that we design the augmentation so that we can adjust the dissimilarity measure (i.e., the feature weights $\{\bfgamma^{(w)}, 
 \bflambda^{(w)} \}$ in Eq.\eqref{eq:Objective})
 by modifying the augmented query vector.
 Since mp-LSH-VA, unfortunately, does not support the cosine-distance, we set $\bfeta^{(w)} = \bfzero$ in this subsection.
 We define the weighted sum query by%
 \footnote{
A semicolon denotes the row-wise concatenation of vectors, like in matlab.
} 
\begin{align}
\overline{\bfq} & =(\overline{\bfq}_1; \cdots; \overline{\bfq}_G) = \textstyle \sum_{w=1}^W \big( \phi_1^{(w)} \bfq_1^{(w)}; \cdots; \phi_G^{(w)} \bfq_G^{(w)} \big),
\notag\\
&  \mbox{where}\quad 
 \phi_g^{(w)}   =  \gamma_g^{(w)} + \lambda_g^{(w)}.
 \notag
\end{align}
 We augment the queries and the samples 
 \ifShortVersion
\else
as follows:
\begin{align}
\widetilde{\bfq} &= (\overline{\bfq}; \bfr),
&
\widetilde{\bfx} &= (\bfx; \bfy),
\notag
\end{align}
\fi
where
$\bfr  \in \mathbb{R}^M$ is a (vector-valued) function of $\{\bfq^{(w)}\}$,
and 
$\bfy \in \mathbb{R}^M$ is a function of $\bfx$.
We constrain the augmentation $\bfy$ for the sample vector so that it satisfies,
for a constant $c_1 \geq 1$,
\begin{align}
\|\widetilde{\bfx}\|_2 &= c_1,
\mbox{ i.e., }
\|\bfy\|_2^2  = c_1^2 -  \|\bfx\|_2^2.
\label{eq:L2NormCondition}
\end{align}
Under this constraint, the norm of any augmented sample is equal to $c_1$,
which allows us to use sign-LSH (Proposition~\ref{prpt:SignLSH}) to perform L2-NNS.
The \LTwoDissimilarity between the query and a sample in the augmented space can be expressed as
\begin{align}
 \pnorm{\widetilde{\bfq} - \widetilde{\bfx}}{2}^2
 &=
 -2\left( \overline{\bfq}^\T {\bfx}  + \bfr^\T \bfy\right)
 + \const
 \label{eq:AugmentedL2Distance}
 \end{align}
  For $M = 1$, only the choice satisfying Eq.\eqref{eq:L2NormCondition} is simple-LSH (for $r = 0$), given in Proposition~\ref{prpt:SimpleLSH}.
 We consider the case for $M \geq 2$, and design $\bfr$ and $\bfy$ so that 
 Eq.\eqref{eq:AugmentedL2Distance} matches the MP dissimilarity \eqref{eq:Objective}.

\def\figsize{0.31\textwidth}
\begin{figure*}[t]
  \centering
  \subfigure[L2NNS ($\gamma = 1, \lambda = 0$)]{
  \includegraphics[width=\figsize]{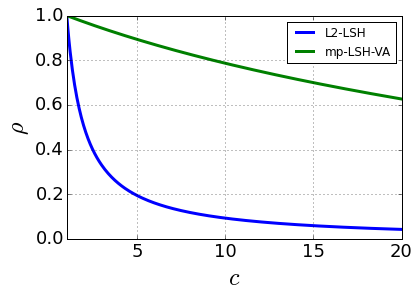}
 }
  \subfigure[MIPS ($\gamma = 0, \lambda = 0$)]{
  \includegraphics[width=\figsize]{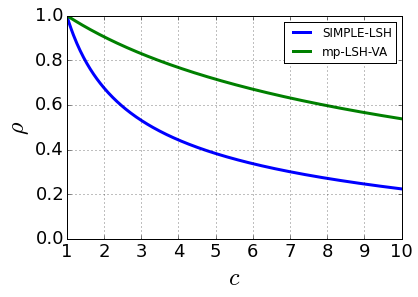}
 }
  \subfigure[Mixed ($\gamma = 0.5, \lambda = 0.5$)]{
  \includegraphics[width=\figsize]{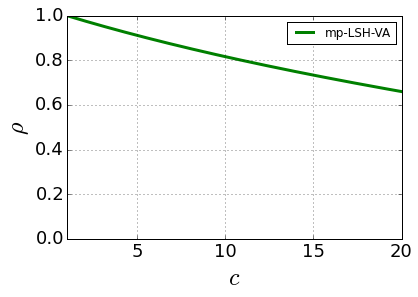}
 }
  \vspace{-2mm}
  \caption{
  Theoretical values $\rho = \frac{\log p_1}{\log p_2}$ (lower is better), which indicates
  the LSH performance 
  (see Proposition~\ref{prpt:CANNS}).
  The horizontal axis indicates $c$ for $c$-ANNS. 
  }
  \label{fig:TheoreticalComparison}
  \vspace{-2mm}
\end{figure*}

 The augmentation that matches the MP dissimilarity is not unique.
 Here, we introduce the following easy construction with $M = G+3$:
\begin{align}
\widetilde{\bfq} &=\textstyle  \Big(\widetilde{\bfq}';
\sqrt{c_2^2 - \|\widetilde{\bfq}'\|_2^2} \Big), \quad
\widetilde{\bfx} = (\widetilde{\bfx}'; 0) \qquad \text{ where}
 \label{eq:MPLSHVA}\\
 \widetilde{\bfq}' & =
\textstyle
\Big(
\underbrace{
\overline{\bfq}_1; \cdots; \overline{\bfq}_G
}_{
\overline{\bfq}
 \in \mathbb{R}^L}
 \; ; \;
\underbrace{\textstyle
\sum_{w=1}^W \gamma_1^{(w)}; \cdots; \sum_{w=1}^W \gamma_G^{(w)}
; 
 0
;
\mu
}_{\bfr' \in \mathbb{R}^{G+2}} \Big),
 \normalsize
 \nonumber\\
\widetilde{\bfx}' &=\textstyle  \Big( 
\underbrace{
 \bfx_1; \cdots ;   \bfx_G 
 }_{\bfx \in \mathbb{R}^L}
\;;\;
\underbrace{\textstyle
 - \frac{  \|\bfx_1\|_2^2}{2}; \cdots;  - \frac{  \|\bfx_{K}\|_2^2}{2}
 ; 
\nu
;
\frac{1}{2}
 }_{\bfy' \in \mathbb{R}^{G+2}}
 \Big).
\notag
 \end{align}
 Here, 
 we defined
 \ifShortVersion
\else
 \begin{align}
\mu & = \textstyle -  \sum_{w=1}^W   \sum_{g=1}^G \gamma_g^{(w)} \| \bfq_g^{(w)}\|_2^2,
 \notag \\
 \nu &=\textstyle \sqrt{c_1^2 -  \left(\| \bfx\|_2^2 + \frac{1}{4}\sum_{g=1}^G\| \bfx_g\|_2^4 + \frac{1}{4}\right)},
 \notag\\
 c_1^2& = \textstyle \max_{\bfx \in \mcX}  \left(\| \bfx\|_2^2 + \frac{1}{4}\sum_{g=1}^G\|\bfx_g\|_2^4 + \frac{1}{4}\right),
\notag\\
c_2^2 & =\textstyle \max_{\bfq} \|\widetilde{\bfq}'\|_2^2.
 \notag
\end{align}
\fi
With the vector augmentation \eqref{eq:MPLSHVA},
Eq.\eqref{eq:AugmentedL2Distance}
matches Eq.\eqref{eq:Objective} up to a constant
\ifWithSectionNumber
(see Appendix~\ref{sec:Proof.MPLSHVA}):
\else
(see Appendix):
\fi
\[\pnorm{\widetilde{\bfq} - \widetilde{\bfx}}{2}^2 = c_1^2+c_2^2 - 2\widetilde{\bfq}^\T \widetilde{\bfx} =  \mcL_{\mathrm{mp}}(\{\bfq^{(w)}\}, \bfx) + \text{const.}\]
The collision probability, i.e., the probability that the query and the sample are given the same code,
 can be analytically computed:
\begin{theorem} 
\label{thrm:MPLSHVA}
Assume that
the samples are rescaled so that $\max_{\bfx \in \mcX} \|\bfx\|_2 \leq 1$ and $\|\bfq^{(w)}\|_2 \leq 1, \forall w$.
For the MP dissimilarity $\mcL_{\mathrm{mp}}(\{\bfq^{(w)}\}, \bfx)$, given by Eq.\eqref{eq:Objective},
with $\bfeta^{(w)} = \bfzero, \forall w$,
the asymmetric hash functions
\begin{align}
h^{\mathrm{VA-q}}_{\bfa}(\{\bfq^{(w)}\})
&= h^{\mathrm{sign}}_{\bfa}(\widetilde{\bfq})
 = \mathrm{sign}(\bfa^\T \widetilde{\bfq}),
\notag\\
h^{\mathrm{VA-x}}_{\bfa}(\bfx)
&= h^{\mathrm{sign}}_{\bfa}(\widetilde{\bfx})
 = \mathrm{sign}(\bfa^\T \widetilde{\bfx}),
\notag
\end{align}
where $\widetilde{\bfq}$ and $\widetilde{\bfx}$ are given by
Eq.\eqref{eq:MPLSHVA},
satisfy
\begin{align}
\textstyle
& \mathbb{P} \big(h_{\bfa}^{\mathrm{VA-q}} ( \! \{\bfq^{(w)}\} \!) =\! h_{\bfa}^{\mathrm{VA-x}} (\bfx) \big)
\notag\\
& \qquad
= \textstyle F^{\mathrm{sign}}\left(1 + \frac{\mcL_{\mathrm{mp}}(\{\bfq^{(w)}\}, \bfx) - 2\pnorm{\bflambda}{1}}{2 c_1 c_2} \right).
\notag
\end{align}
\end{theorem}
(Proof)
Via construction, it holds that $\|\widetilde{\bfx}\|_2 = c_1$ and $\|\widetilde{\bfq}\|_2 = c_2$,
 and simple calculations 
 \ifWithSectionNumber
(see Appendix~\ref{sec:Proof.MPLSHVA}) 
\else
(see Appendix) 
\fi
give $\widetilde{\bfq}^\T \widetilde{\bfx} = \pnorm{\bflambda}{1} - \frac{\mcL_{\mathrm{mp}}(\{\bfq^{(w)}\}, \bfx)}{2}$.
 Then, applying Propostion~\ref{prpt:SignLSH} immediately proves the theorem. 
\QED

Figure~\ref{fig:TheoreticalComparison} depicts
the theoretical value of $\rho = \frac{\log p_1}{\log p_2}$
of mp-LSH-VA, computed by using Thoerem~\ref{thrm:MPLSHVA}, 
\ifShortVersion
\else
for different weight settings for $G = 1$.
Note that $\rho$
determines the quality of LSH (smaller is better)
for $c$-ANNS performance
(see Proposition~\ref{prpt:CANNS}).
In the case for L2-NNS and MIPS,
the $\rho$ values of the standard LSH methods, i.e.,
L2-LSH (Proposition~\ref{prpt:L2LSH}) and simple-LSH (Proposition~\ref{prpt:SimpleLSH}),
are also shown for comparison.

\fi

Although mp-LSH-VA offers attractive flexibility with adjustable dissimilarity,
Figure~\ref{fig:TheoreticalComparison} implies its inferior performance to the standard methods,
especially in the L2-NNS case.
\ifShortVersion
\else
The reason might be a too strong asymmetry between the query and the samples:
a query and a sample are far apart in the augmented space, even if they are close to each other in the original space.
We can see this from the first $G$ entries in $\bfr$ and $\bfy$ in Eq.\eqref{eq:MPLSHVA},
respectively.
Those entries for the query are non-negative, i.e., $r_m \geq 0$ for $m = 1, \ldots, G$,
while the corresponding entries for the sample are non-positive, i.e., $y_m \leq 0$ for $m = 1, \ldots, G$.
\fi
We believe that there is room to improve the performance of mp-LSH-VA, 
e.g., by adding constants and changing the scales of some augmented entries,
which we leave as our future work.

In the next subsections,
we propose alternative approaches, where
codes are as symmetric as possible,
and down-weighting is done by changing the metric in the code space.
This effectively keeps close points in the original space close in the code space.

\ifShortVersion
\else
\subsection{Multiple purpose LSH with Code Concatenation (mp-LSH-CC)}
\fi
\label{sec:mp-LSH-CC}

Let $\overline{\gamma}_g =  \sum_{w=1}^W \gamma_g^{(w)} $,
$\overline{\eta}_g =  \sum_{w=1}^W \eta_g^{(w)} $,
and
$\overline{\lambda}_g =  \sum_{w=1}^W \lambda_g^{(w)} $,
and define the \emph{metric-wise} weighted average queries by
$\overline{\bfq}_g^{\mathrm{L2}} = \frac{\sum_{w=1}^W \gamma_g^{(w)} \bfq_g^{(w)}}{\overline{\gamma}_g}$, 
$\overline{\bfq}_g^{\mathrm{cos}} = \sum_{w=1}^W \eta_g^{(w)} \frac{\bfq_g^{(w)}}{\|\bfq_g^{(w)}\|_2}$,
and
$\overline{\bfq}_g^{\mathrm{ip}} = \sum_{w=1}^W \lambda_g^{(w)} \bfq_g^{(w)}$.

Our second proposal, 
called multiple purpose LSH with code concatenation (mp-LSH-CC),
simply concatenates multiple LSH codes,
and performs NNS under the following distance metric at query time:
%
%
\begin{align}
\!\! \!
\mcD_{\mathrm{CC}}(\{\bfq^{(w)}\}, {\bfx})
&\!= \!
\sum_{g=1}^G \!
\sum_{t=1}^T  \!
\textstyle
\Big(
 \overline{\gamma}_g R\sqrt{\frac{\pi}{2}} \left|h_{t}^{\mathrm{L2}}(\overline{\bfq}_g^{\mathrm{L2}}) \!-\! h_{t}^{\mathrm{L2}}({\bfx}_g) \right|
\notag\\
& \hspace{-5mm}
\textstyle
+
\pnorm{\overline{\bfq}_g^{\mathrm{cos}}}{2} \left|h_{t}^{\mathrm{sign}}(\overline{\bfq}_g^{\mathrm{cos}}) - h_{t}^{\mathrm{sign}}({\bfx}_g) \right|
\notag\\
& \hspace{-5mm}
\textstyle
+\pnorm{\overline{\bfq}_g^{\mathrm{ip}}}{2}  \left|h_{t}^{\mathrm{smp-q}}(\overline{\bfq}_g^{\mathrm{ip}}) - h_{t}^{\mathrm{smp-x}}({\bfx}_g) \right|\Big)
\label{eq:CodeDistanceCC},
\end{align}
where $h_{t}^{\mbox{---}}$ denotes the $t$-th independent draw of the corresponding LSH code
for $t = 1, \ldots, T$.
The distance \eqref{eq:CodeDistanceCC}
is a \emph{multi-metric}, a linear combination of metrics \cite{Bustos12},
in the code space.
 For a multi-metric,
we can use the \emph{cover tree} \cite{Beygelzimer06}
for efficient (exact) NNS.
Assuming that all adjustable linear weights are upper-bounded by 1,
the cover tree expresses neighboring relation between samples,
taking all possible weight settings into account.
NNS is conducted by bounding the code metric for a given weight setting.
Thus, mp-LSH-CC allows selective exploration of hash buckets, so that
we only need to accurately measure the distance to the samples
assigned to the hash buckets within a small code distance.
The query time complexity of the cover tree is $O(\kappa^{12} \log N)$, 
where $\kappa$ is a data-dependent \emph{expansion constant} 
\cite{Heinonen01}.
\ifShortVersion
\else
Another good aspect of the cover tree is that it allows dynamic insertion and deletion of new samples,
and therefore, it lends itself naturally to the streaming setting.
\fi
\ifWithSectionNumber
Appendix~\ref{sec:CoverTreeDetail} describes further details.
\else
Further details are described in Appendix.
\fi

In the pure case for L2, cosine, or IP dissimilarity,
the hash code of mp-LSH-CC is equivalent to the base LSH code,
and therefore, the performance is guaranteed by Propositions~\ref{prpt:L2LSH}--\ref{prpt:SimpleLSH}, respectively.
However, mp-LSH-CC is not optimal in terms of memory consumption and NNS efficiency.
This inefficiency comes from the fact that it \emph{redundantly} stores the same angular (or cosine-distance) information
into each of the L2-, sign-, and simple-LSH codes.
Note that the information of a vector is dominated by its angular components unless the dimensionality $L$ is very small.

\ifShortVersion
\else
\subsection{Multiple purpose LSH with Code Augmentation and Transformation (mp-LSH-CAT)}
\fi
\label{sec:mp-LSH-CAT}

Our third proposal, called multiple purpose LSH with code augmentation and transformation (mp-LSH-CAT),
offers significantly less memory requirement and faster NNS
than mp-LSH-CC
by sharing the angular information for all considered dissimilarity measures.
 Let
 \[ 
 \overline{\bfq}_g^{\mathrm{L2+ip}} 
 =\textstyle  \sum_{w=1}^W (\gamma_g^{(w)} + \lambda_g^{(w)}) \bfq_g^{(w)}.
 \]
We essentially use sign-hash functions that we augment with norm information of the data, giving us the following augmented codes:
\begin{align}
&\bfH^{\mathrm{CAT-q}}(\{\bfq^{(w)}\})
= \left(\bfH(\overline{\bfq}^{\mathrm{L2+ip}}) ; \bfH(\overline{\bfq}^{\mathrm{cos}}) ; \textbf{0}_G^\top \right)\mbox{ and}
\label{eq:mpLSHCATCodeQ}\\
&\bfH^{\mathrm{CAT-x}}(\bfx) 
 = \Big(\widetilde{\bfH}(\bfx) ; \bfH(\bfx) ; \bfj^\T(\bfx)\Big)\mbox{, where}
 \label{eq:mpLSHCATCodeX}\\
 &\bfH(\bfv)
 =\Big(\mathrm{sign} (\bfA_1 \bfv_1), \ldots, \mathrm{sign} (\bfA_G \bfv_G) \Big),
  \label{eq:mpLSHCATSignLSHPart}\\
 &\widetilde{\bfH}(\bfv)
= \Big(\pnorm{\bfv_1}{2}\mathrm{sign} (\bfA_1 \bfv_1), \ldots, \pnorm{\bfv_G}{2}\mathrm{sign} (\bfA_G \bfv_G) \Big),\notag\\
&\bfj(\bfv) = \Big(\pnorm{\bfv_1}{2}^2; \ldots; \pnorm{\bfv_G}{2}^2\Big) ,\notag
\normalsize
\end{align}
for a partitioned vector $ \bfv = (\bfv_1; \ldots; \bfv_G)\in\mathbb{R}^L$ and 
$\textbf{0}_G = (0; \cdots; 0) \in \mathbb{R}^G$.
Here, each entry of $\bfA = (\bfA_1, \ldots, \bfA_G) \in \mathbb{R}^{T \times L}$ follows $A_{t, l}  \sim \mathcal{N}(0, 1^2)$.

For two matrices $\bfH', \bfH'' \in \mathbb{R}^{(2T+1)\times G}$ 
in the transformed hash code space, 
we measure the distance with the following multi-metric:
\begin{align}
\mcD_{\mathrm{CAT}}(\bfH', \bfH'')
&= 
\textstyle \sum_{g=1}^G
\bigg(
\alpha_g
\sum_{t=1}^{T}  
  \left| H_{t, g}' - {H}_{t, g}'' \right|\notag\\
\textstyle
&\hspace{-25mm}
\textstyle
+\beta_g
\sum_{t=T+1}^{2T}  
  \left|H_{t, g}' - {H}_{t, g}'' \right|
+ \overline{\gamma}_g \frac{T}{2} \left| H_{2T+1, g}' - {H}_{2T+1, g}'' \right|
  \bigg),\label{eq:CodeDistanceCAT}
\end{align}
where
$\alpha_g = \textstyle 
\pnorm{\overline{\bfq}_g^{\mathrm{L2+ip}}}{2}$
and
$\beta_g = \pnorm{\overline{\bfq}_g^{\mathrm{cos}}}{2}$.

Although the hash codes consist of $(2T+1)G$ entries,
we do not need to store all the entries,
and computation can be simpler and faster by first computing the total number of collisions 
in the sign-LSH part \eqref{eq:mpLSHCATSignLSHPart} for $g = 1, \ldots, G$:
\begin{align}
\mcC_g&(\bfv', \bfv'') = \textstyle \sum_{t=1}^T \Big\{ \big(\bfH (\bfv') \big)_{t, g}  = \big( \bfH ({\bfv''}) \big)_{t, g}\Big\}.
\label{eq:numberCollisions}
\end{align}
Note that this computation, which dominates the computation cost for evaluating code distances,
can be performed efficiently with bit operations.
With the total number of collisions \eqref{eq:numberCollisions},
the metric \eqref{eq:CodeDistanceCAT} between a query set $\{\bfq^{(w)}\}$ and a sample $\bfx$
can be expressed as
\begin{align}
\!\!\!\mcD_{\mathrm{CAT}}& \Big(\bfH^{\mathrm{CAT-q}}(\{\bfq^{(w)}\}), \bfH^{\mathrm{CAT-x}}(\bfx) \Big)
\textstyle\notag\\
 = &\textstyle \sum_{g=1}^G
\bigg(\alpha_g\Big(T + \pnorm{\bfx_g}{2}\big( T - 2\mcC_g(\overline{\bfq}^{\mathrm{L2+ip}}, \bfx)\big)\Big)\notag\\
\textstyle
&+2\beta_g\big(T - \mcC_g(\overline{\bfq}^{\mathrm{cos}}, \bfx)\big)
+ \overline{\gamma}_g \textstyle\frac{T}{2}\pnorm{\bfx_g}{2}^2
  \bigg).
 \label{eq:queryToDataDistanceCAT}
\end{align}
Given a query set, this can be computed from $\bfH(\bfx) \in \mathbb{R}^{T \times G}$ and $\pnorm{\bfx_g}{2}$ for $g = 1, \ldots, G$.
Therefore, we only need to store the pure $TG$ sign-bits, which is required by sign-LSH alone, 
and $G$ additional float numbers.

Similarly to mpLSH-CC, we use the cover tree for efficient NNS based on the code distance \eqref{eq:CodeDistanceCAT}.
In the cover tree construction, we set the metric weights to their upper-bounds, i.e., $\alpha_g = \beta_g = \overline{\gamma}_g = 1$,
and measure the distance between samples by
\begin{align}
\!\!\!\mcD_{\mathrm{CAT}}& \Big(\bfH^{\mathrm{CAT-x}}(\bfx') , \bfH^{\mathrm{CAT-x}}(\bfx'') \Big)
\notag\\
= & \textstyle \sum_{g=1}^G
\bigg(
  \left|\pnorm{\bfx_g'}{2} - \pnorm{{\bfx}''_g}{2} \right|\mcC_g(\bfx', {\bfx''})\notag\\
  &+ (\pnorm{\bfx_g'}{2} + \pnorm{{\bfx}''_g}{2} + 2)\big(T - \mcC_g(\bfx', {\bfx''})\big)\notag\\
\textstyle
&
\textstyle
+\frac{T}{2} \left| \pnorm{\bfx_g'}{2}^2 - \pnorm{{\bfx}''_g}{2}^2 \right|
  \bigg).
 \label{eq:dataToDataDistanceCAT}
\end{align}

 
Since the collision probability
can be zero,
we cannot directly apply the standard LSH theory with the $\rho$ value guaranteeing the ANNS performance.
Instead, we show that the metric \eqref{eq:CodeDistanceCAT} of mpLSH-CAT approximates the MP dissimilarity  \eqref{eq:Objective},
and the quality of ANNS is guaranteed.

\begin{theorem}
\label{thrm:CATL2PlusIPApproximation}
For $\bfeta^{(w)} = \bfzero, \forall w$, it is
\[
\textstyle
\lim_{T \to \infty} \frac{\mcD_{\mathrm{CAT}} }{T} =
\frac{1}{2}\mcL_{\mathrm{mp}}(\{\bfq^{(w)}\}, \bfx) + \text{const.} + \text{error},
\]
with $|\text{error}| \leq 0.2105\left(\pnorm{\bflambda}{1} + \pnorm{\bfgamma}{1}\right).$
\end{theorem}
(proof is given in Appendix~\ref{sec:Proof.CATL2PlusIPApproximation}).

%

\begin{theorem}
\label{thrm:CATcosineApproximation}
For $\bfgamma^{(w)} = \bflambda^{(w)} = \bfzero, \forall w$, it is
\[
\textstyle
\lim_{T \to \infty} \frac{\mcD_{\mathrm{CAT}} }{T} =
\frac{1}{2}\mcL_{\mathrm{mp}}(\{\bfq^{(w)}\}, \bfx) + \text{const.} + \text{error},
\]
with $\text{error}| \leq 0.2105\pnorm{\bfeta}{1}$.
\end{theorem}
(proof is given in Appendix~\ref{sec:Proof.CATcosineApproximation}).
\begin{corollary}
\label{cor:CATApproximation}
 \[
 \textstyle
2\lim_{T \to \infty} \frac{\mcD_{\mathrm{CAT}} }{T} =
\mcL_{\mathrm{mp}}(\{\bfq^{(w)}\}, \bfx) + \text{const.} + \text{error},
\]
with
\[|\text{error}| \leq 0.421.\]
\end{corollary}
The error with a maximum of $0.421$ ranges one order of magnitude below the MP dissimilarity having itself a range of $4$.
Note that Corollary~\ref{cor:CATApproximation} implies good approximation for the boundary cases, squared-L2-, IP- and cosine-distance, through mpLSH-CAT
since they are special cases of weights: For example, mpLSH-CAT approximates \LTwoDissimilarity when setting $\bflambda^{(w)} = \bfeta^{(w)} = \bfzero, \forall w$.
%
The following theorem guarantees ANNS to succeed with mp-LSH-CAT for pure MIPS case with specified probability (proof is given in Appendix~\ref{sec:Proof.CATNNSGuarantee}):
\begin{theorem}
\label{thrm:CATNNSGuarantee}
Let $S_0 \in (0,2)$, $cS_0 \in (S_0 + 0.2105,2)$ and set \[T \geq \textstyle \frac{48}{(t_2-t_1)^2}\log(\frac{n}{\varepsilon}),\] where $t_2 > t_1$ depend on $S_0$ and $c$ (see Appendix~\ref{sec:Proof.CATNNSGuarantee} for details).
With probability larger than $1 - \varepsilon - \left(\frac{\varepsilon}{n}\right)^\frac{3}{2}$ mp-LSH-CAT guarantees $c$-ANNS with respect to $\mcL_{\mathrm{ip}}$ (MIPS). 
\end{theorem}
Note that it is straight forward to show Theorem~\ref{thrm:CATNNSGuarantee} for squared-L2- and cosine-distance.

In  Section~\ref{sec:Experiment}, 
we will empirically show the good performance of mpLSH-CAT in general cases.

 \def\figsize{0.25\textwidth}
\begin{figure*}[t]
\ifGraphWithoutSign
  \centering
  \subfigure[L2NNS ($\gamma = 1, \lambda = 0$)]{
  \includegraphics[width=\figsize]{ML/w_0p0_1p0/t5_h256_wosign.png} 
 }
 \qquad
  \subfigure[MIPS ($\gamma = 0, \lambda = 1$)]{
  \includegraphics[width=\figsize]{ML/w_1p0_0p0/t5_h256_wosign.png} 
 }
 \qquad
  \subfigure[Mixed ($\gamma = 0.5, \lambda = 0.5$)]{
  \includegraphics[width=\figsize]{ML/w_0p5_0p5/t5_h256_wosign.png} 
 }
\else
 \centering
  \subfigure[L2NNS ($\gamma = 1, \lambda = 0$)]{
  \includegraphics[width=\figsize]{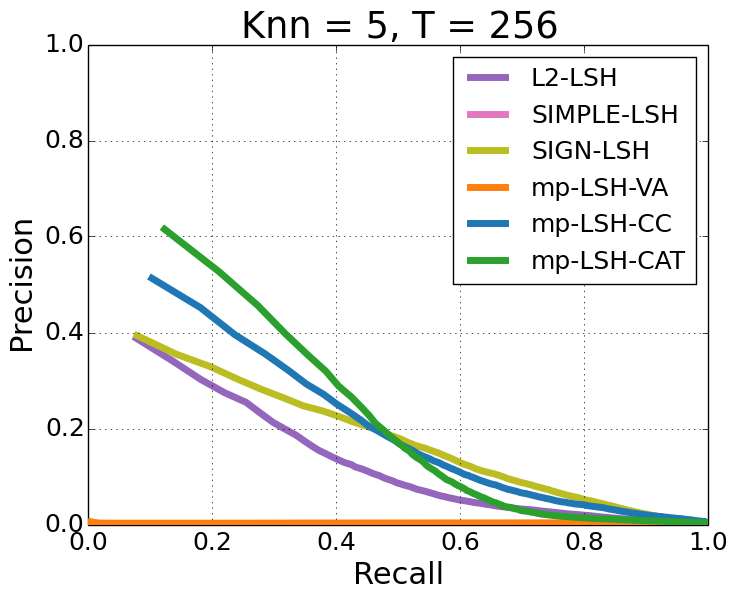} 
 }
 \qquad
  \subfigure[MIPS ($\gamma = 0, \lambda = 1$)]{
  \includegraphics[width=\figsize]{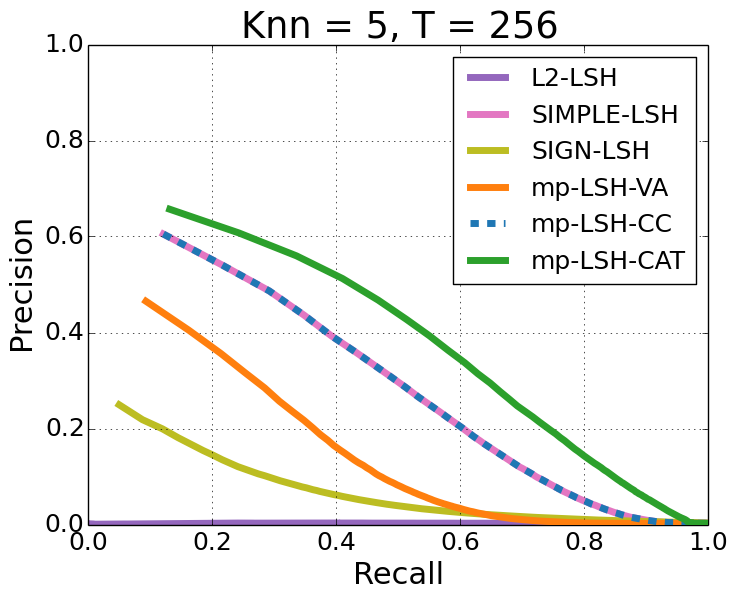} 
 }
 \qquad
  \subfigure[Mixed ($\gamma = 0.5, \lambda = 0.5$)]{
  \includegraphics[width=\figsize]{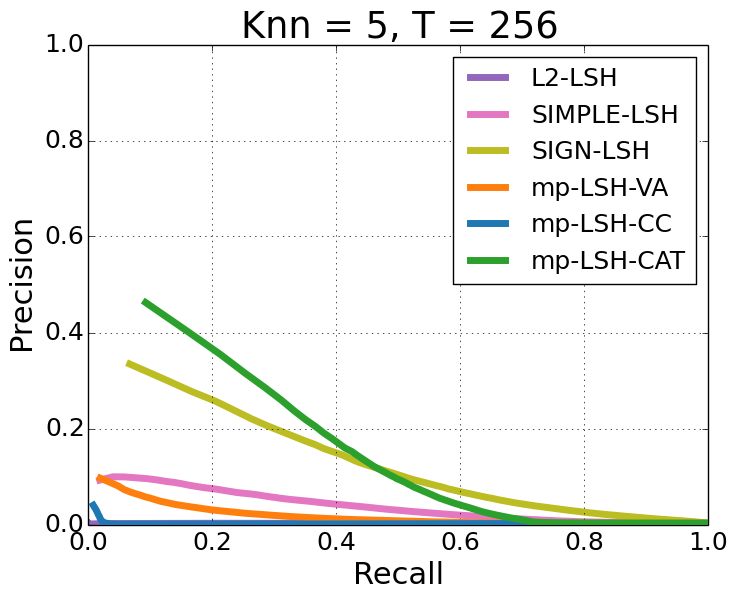} 
 }
 \fi
  \vspace{-2mm}
  \caption{ 
 Precision recall curves (higher is better) on MovieLens10M data for $K= 5$ and $T = 256$.
 }
  \label{fig:ExperimentCFML}
  \vspace{-2mm}
\end{figure*}

 \def\figsize{0.25\textwidth}
\begin{figure*}[t]
\ifGraphWithoutSign
  \centering
  \subfigure[L2NNS ($\gamma = 1, \lambda = 0$)]{
  \includegraphics[width=\figsize]{NF/w_0p0_1p0/t5_h256_wosign.png} 
 }
 \qquad
  \subfigure[MIPS ($\gamma = 0, \lambda = 1$)]{
  \includegraphics[width=\figsize]{NF/w_1p0_0p0/t5_h256_wosign.png} 
 }
 \qquad
  \subfigure[Mixed ($\gamma = 0.5, \lambda = 0.5$)]{
  \includegraphics[width=\figsize]{NF/w_0p5_0p5/t5_h256_wosign.png} 
 }
\else
 \centering
  \subfigure[L2NNS ($\gamma = 1, \lambda = 0$)]{
  \includegraphics[width=\figsize]{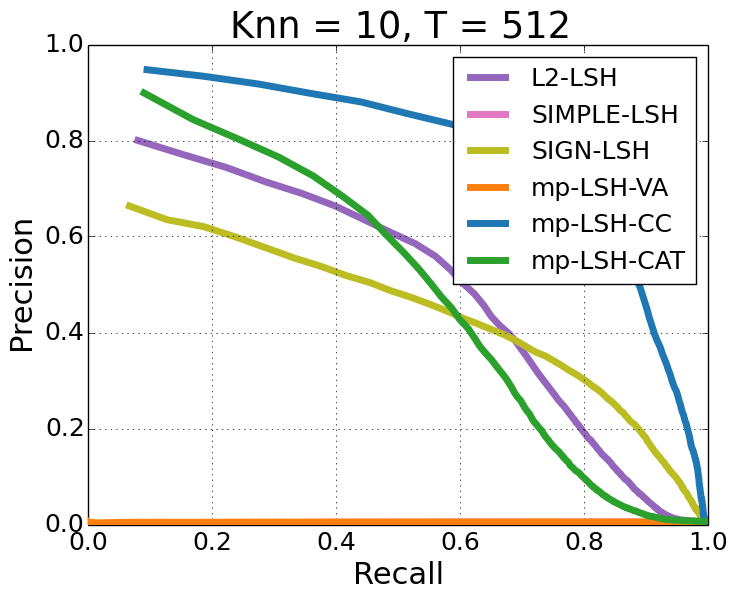} 
 }
 \qquad
  \subfigure[MIPS ($\gamma = 0, \lambda = 1$)]{
  \includegraphics[width=\figsize]{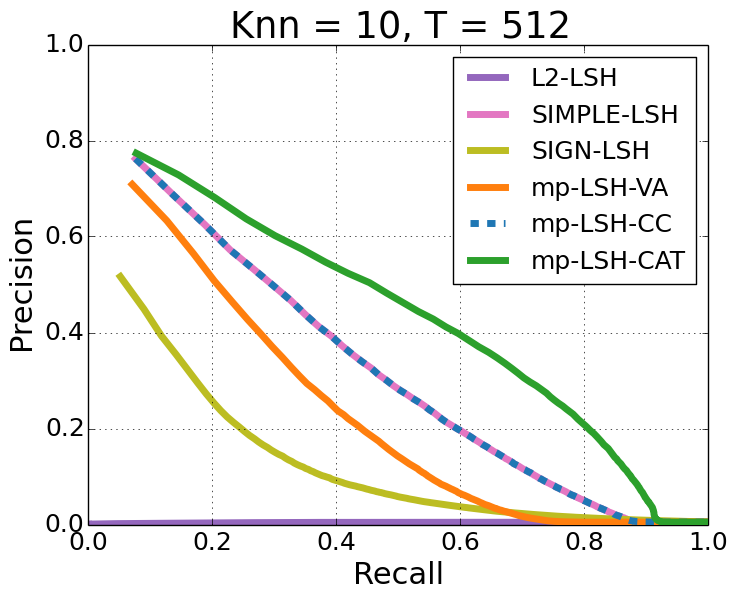} 
 }
 \qquad
  \subfigure[Mixed ($\gamma = 0.5, \lambda = 0.5$)]{
  \includegraphics[width=\figsize]{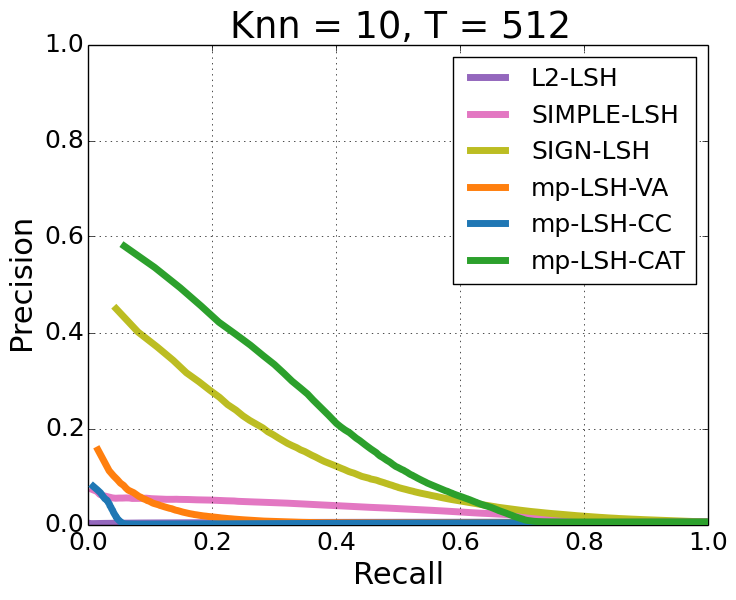} 
 }
 \fi
  \vspace{-2mm}
  \caption{ 
 Precision recall curves on NetFlix data for $K= 10$ and $T = 512$.
 }
  \label{fig:ExperimentCFNF}
  \vspace{-2mm}
\end{figure*}

\subsection{Memory Requirement}
\label{sec:Memoery}
For all LSH schemes, one can trade off the memory consumption and accuracy performance by changing the hash bit length $T$.
%
However, the memory consumption for specific hashing schemes heavily differs from the other schemes such that
a comparison of performance is inadequate for a globally shared $T$.
In this subsection, we
derive individual numbers of hashes for each scheme, given a fixed memory budget.

We count the theoretically minimal number of bits required to store the hash code of one data point.
The two fundamental components we are confronted with are sign-hashes and discretized reals.
Sign-hashes can be represented by exactly one bit. For the reals we choose a resolution
such that their discretizations take values in a set of fixed size.
The L2-hash function $h_{\bfa, b}^{\mathrm{L2}} (\bfx) 
= \textstyle
 \left\lfloor R^{-1}(\bfa^\T \bfx + b) \right\rfloor$
is a random variable with potentially infinite, discrete values.
Nevertheless we can come up with a realistic upper-bound of values the L2-hash essentially takes.
Note that $R^{-1}(\bfa^\T \bfx)$ follows a $\mathcal{N}(\mu=0, \sigma = (R\pnorm{x}{2})^{-1})$ distribution and $\pnorm{x}{2} \leq 1$.
Then $\mathbb{P}(|R^{-1}(\bfa^\T \bfx)| > 4\sigma) < 10^{-4}$. Therefore L2-hash essentially takes one of $\frac{8}{R}$ discrete values
stored by $3-\log_2(R)$ bits. Namely, for $R = 2^{-10} \approx 0.001$ L2-hash requires 13 bits.
We also store the norm-part of mp-LSH-CAT using 13 bits.

Denote by $\storeCat(T)$ the required storage of mp-LSH-CAT. Then
$\storeCat(T) = T_\mathrm{CAT} + 13$, which we set as our fixed memory budget for a given $T_\mathrm{CAT}$.
The baselines sign- and simple-LSH, so mp-LSH-VA are pure sign-hashes, thus giving them a budget of $T_{\mathrm{sign}} = T_{\mathrm{smp}} = T_{\mathrm{VA}} = \storeCat(T)$ hashes.
As discussed above L2-LSH may take $T_{\mathrm{L2}} = \frac{\storeCat(T)}{13}$ hashes.
For mp-LSH-CC we allocate a third of the budget for each of the three components giving
$\bfT_{\mathrm{CC}} = (T_{\mathrm{CC}}^{\mathrm{L2}}, T_{\mathrm{CC}}^{\mathrm{sign}}, T_{\mathrm{CC}}^{\mathrm{smp}}) = \storeCat(T) \cdot (\frac{1}{39},\frac{1}{3},\frac{1}{3})$.
This consideration is used when we compare mp-LSH-CC and mp-LSH-CAT
in Section~\ref{sec:Experiment.TexMex}.

\begin{table*}[t]
\centering
\caption{ANNS Results for mp-LSH-CC with $\bfT_{\mathrm{CC}} = (T_{\mathrm{CC}}^{\mathrm{L2}}, T_{\mathrm{CC}}^{\mathrm{sign}}, T_{\mathrm{CC}}^{\mathrm{smp}})=(1024, 1024, 1024)$.}
\label{tb:ResultCC}
\begin{tabular}{@{}lllllllll@{}}
\toprule
        & \multicolumn{3}{l}{Recall@k} & \multicolumn{3}{l}{Query time (msec)} & \multirow{2}{*}{\begin{tabular}[c]{@{}l@{}}Cover Tree\\ Construction (sec) \end{tabular}} & \multirow{2}{*}{\begin{tabular}[c]{@{}l@{}}Storage \\ per sample \end{tabular}} \\
        & 1         & 5         & 10       & 1           & 5           & 10          &          &  \\ \midrule

L2 & 0.53 & 0.76 & 0.82 & 2633.83 & 2824.06 & 2867.00 & 31351  & 4344 bytes \\
MIPS & 0.69 & 0.77 & 0.82 & 3243.51 & 3323.20&  3340.36 & 31351  & 4344 bytes \\
L2+MIPS (.5,.5) & 0.29 & 0.50 & 0.60 & 3553.63 & 3118.93&  3151.44 & 31351  & 4344 bytes \\\bottomrule
\end{tabular}
\vspace{2mm}
\centering
\caption{ANNS Results with mp-LSH-CAT with $T_{\mathrm{CAT}}=1024$.
}
\label{tb:ResultCAT}
\begin{tabular}{@{}lllllllll@{}}
\toprule
        & \multicolumn{3}{l}{Recall@k} & \multicolumn{3}{l}{Query time (msec)} & \multirow{2}{*}{\begin{tabular}[c]{@{}l@{}}Cover Tree\\ Construction (sec) \end{tabular}} & \multirow{2}{*}{\begin{tabular}[c]{@{}l@{}}Storage \\ per sample \end{tabular}} \\
        & 1         & 5         & 10       & 1           & 5           & 10          &          &  \\ \midrule
L2 & 0.52 & 0.80 & 0.89 & 583.85 & 617.02 & 626.02 & 41958  & 224 bytes \\
MIPS & 0.64 & 0.76 & 0.85 & 593.11 & 635.72 & 645.14 & 41958  & 224 bytes \\
L2+MIPS (.5,.5) & 0.29 & 0.52 & 0.62 &476.62 & 505.63 & 515.77 & 41958  & 224 bytes \\\bottomrule
\end{tabular}
\vspace{2mm}
\centering
\caption{ANNS Results for mp-LSH-CC with $\bfT_{\mathrm{CC}} = (T_{\mathrm{CC}}^{\mathrm{L2}}, T_{\mathrm{CC}}^{\mathrm{sign}}, T_{\mathrm{CC}}^{\mathrm{smp}}) = (27, 346, 346)$.
}
\label{tb:ResultCCSameMemory}
\begin{tabular}{@{}lllllllll@{}}
\toprule
        & \multicolumn{3}{l}{Recall@k} & \multicolumn{3}{l}{Query time (msec)} & \multirow{2}{*}{\begin{tabular}[c]{@{}l@{}}Cover Tree\\ Construction (sec) \end{tabular}} & \multirow{2}{*}{\begin{tabular}[c]{@{}l@{}}Storage \\ per sample \end{tabular}} \\
        & 1         & 5         & 10       & 1           & 5           & 10          &          &  \\ \midrule

L2 & 0.35 & 0.49 & 0.59 & 1069.29 & 1068.97 & 1074.40 & 4244  & 280 bytes \\
MIPS & 0.32 & 0.56 & 0.56 & 363.61 & 434.49 & 453.35 & 4244  & 280 bytes \\
L2+MIPS (.5,.5) & 0.04 & 0.07 & 0.08 & 811.72 & 839.91 & 847.35 & 4244  & 280 bytes \\\bottomrule
\end{tabular}
\vspace{-2mm}
\end{table*}

\section{Experiment}
\label{sec:Experiment}

\ifShortVersion
\else
Here, we conduct an empirical evaluation on several real-world data sets.
\fi

\subsection{Collaborative Filtering}
\label{sec:Experiment.CF}

We first evaluate our methods on 
collaborative filtering data,
the MovieLens10M%
\footnote{\url{http://www.grouplens.org/}}
and the Netflix datasets \citep{Funk06}.
Following the experiment in \citep{Shrivastava14,Shrivastava15},
we applied PureSVD \citep{Cremonesi10} to get $L$-dimensional user and item vectors,
where $L = 150$ for MovieLens and $L = 300$ for Netflix.
We centered the samples so that $\sum_{\bfx \in \mcX} \bfx = \bfzero$,
which does not affect the L2-NNS as well as the MIPS solution.

Regarding the $L$-dimensional vector as a single feature group ($G=1$),
we evaluated the performance in L2-NNS ($W = 1, \gamma = 1, \eta = \lambda = 0$),
MIPS ($W = 1, \gamma = \eta = 0, \lambda = 1$),
and their weighted sum 
($W = 2, \gamma^{(1)} = 0.5, \lambda^{(2)} =  0.5,  \gamma^{(2)} = \lambda^{(1)} = \eta^{(1)} = \eta^{(2)} = 0$).
The queries for L2-NNS were chosen randomly from the items,
while the queries for MIPS were chosen from the users.
For each query,
we found its $K= 1, 5, 10$ nearest neighbors in terms of the MP dissimilarity \eqref{eq:Objective} by linear search,
and used them as the ground truth.
We set the hash bit length to $T = 128, 256, 512$,
and rank the samples (items) based on the Hamming distance
 for the baseline methods and mp-LSH-VA.
For mp-LSH-CC and mp-LSH-CAT, we rank the samples based on their code distances \eqref{eq:CodeDistanceCC} and \eqref{eq:CodeDistanceCAT},
respectively.
After that, we drew the precision-recall curve, defined as
\Dummytrue
\ifDummy
$\mathrm{Precision} = \textstyle \frac{\mathrm{relevant seen}}{k}$
and 
$\mathrm{Recall} = \textstyle \frac{\mathrm{relevant seen}}{K}$
\else
\begin{align}
\mathrm{Precision} =& \textstyle \frac{\mathrm{relevant seen}}{k},
&
\mathrm{Recall} =& \textstyle \frac{\mathrm{relevant seen}}{K},
\label{eq:PrecisionRecall}
\end{align}
\fi
for different $k$,
where ``relevant seen'' is the number of the true $K$ nearest neighbors that are ranked within the top $k$ positions by the LSH methods.
Figures~\ref{fig:ExperimentCFML} and \ref{fig:ExperimentCFNF} show the results on MovieLens10M for $K=5$ and $T = 256$ and NetFlix for $K=10$ and $T = 512$, respectively,
where
each curve was averaged over 2000 randomly chosen queries.

We observe that mp-LSH-VA performs very poorly in L2-NNS (as bad as simple-LSH, which is not designed for L2-distance),
although it performs reasonably in MIPS.
On the other hand, mp-LSH-CC and mp-LSH-CAT perform well for all cases.
Similar tendency was observed for other values of $K$ and $T$.
Since poor performance of mp-LSH-VA was shown in theory (Figure~\ref{fig:TheoreticalComparison})
and experiment (Figures~\ref{fig:ExperimentCFML} and \ref{fig:ExperimentCFNF}),
we will focus on mp-LSH-CC and mp-LSH-CAT in the subsequent subsections.

\subsection{Computation Time in Query Search}
\label{sec:Experiment.TexMex}

Next, we evaluate query search time and memory consumption of mp-LSH-CC and mp-LSH-CAT 
on the texmex dataset\footnote{
\url{http://corpus-texmex.irisa.fr/}}  \cite{Jegou11}, which was generated from millions of images by applying the standard SIFT descriptor \cite{Lowe04} with $L=128$.
Similarly to 
\ifWithSectionNumber
Section~\ref{sec:Experiment.CF},
\else
the previous section,
\fi
we conducted experiment on L2-NNS, MIPS, and their weighted sum with the same setting for the weights $\bfgamma, \bfeta, \bflambda$.
\ifNotFullExperiment
We constructed the cover tree with $N = 10^7$ samples, randomly chosen  
from the ANN\_SIFT1B dataset.
The queries were chosen from the defined \emph{query set}, 
and the query for MIPS is normalized so that $\|\bfq\|_2 = 1$.
\fi

We ran the performance experiment on a machine with 48 cores (4 AMD Opteron\texttrademark   6238 Processors) and 512 GB main memory on Ubuntu 12.04.5 LTS.
Tables~\ref{tb:ResultCC}--\ref{tb:ResultCCSameMemory} summarize
\ifShortVersion
\else
recall@$k$, query time, cover tree construction time, and required memory storage.
Here, recall@$k$ is the recall for $K = 1$ and given $k$.
All reported values, except the cover tree construction time, 
\ifNotFullExperiment
are averaged over 100 queries.
\else
are averaged over the queries.
\fi
\fi

We see that mp-LSH-CC (Table~\ref{tb:ResultCC})
and mp-LSH-CAT (Table~\ref{tb:ResultCAT}) for $T = 1024$
perform comparably well in terms of accuracy (see the columns for recall@k).
But mp-LSH-CAT is much faster (see query time) and requires significantly less memory (see storage per sample).
Table~\ref{tb:ResultCCSameMemory} shows the performance of mp-LSH-CC with equal memory requirement to mp-LSH-CAT for $T=1024$.
More specifically, we use different bit length for each dissimilarity measure, and set them to
 $\bfT_{\mathrm{CC}} = (T_{\mathrm{CC}}^{\mathrm{L2}}, T_{\mathrm{CC}}^{\mathrm{sign}}, T_{\mathrm{CC}}^{\mathrm{smp}}) = (27, 346, 346)$,
 with which the memory budget is shared equally for each dissimilarity measure, according to Section~\ref{sec:Memoery}.
By comparing Table~\ref{tb:ResultCAT} and Table~\ref{tb:ResultCCSameMemory},
we see that mp-LSH-CC for $\bfT_{\mathrm{CC}} = (27, 346, 346)$, which uses similar memory storage per sample,
gives significantly worse recall@k than mp-LSH-CAT for $T=1024$.

Thus, we conclude that 
both mp-LSH-CC and mp-LSH-CAT perform well,
but we recommend the latter for the case of limited memory budget,
or in applications where the query search time is crucial.


%

 \def\figsizeL{0.48\textwidth}
 \def\figsizeR{0.48\textwidth}
\begin{figure*}[t]
  \centering
      \subfigure[Trench coats]{
  \includegraphics[width=\figsizeL]{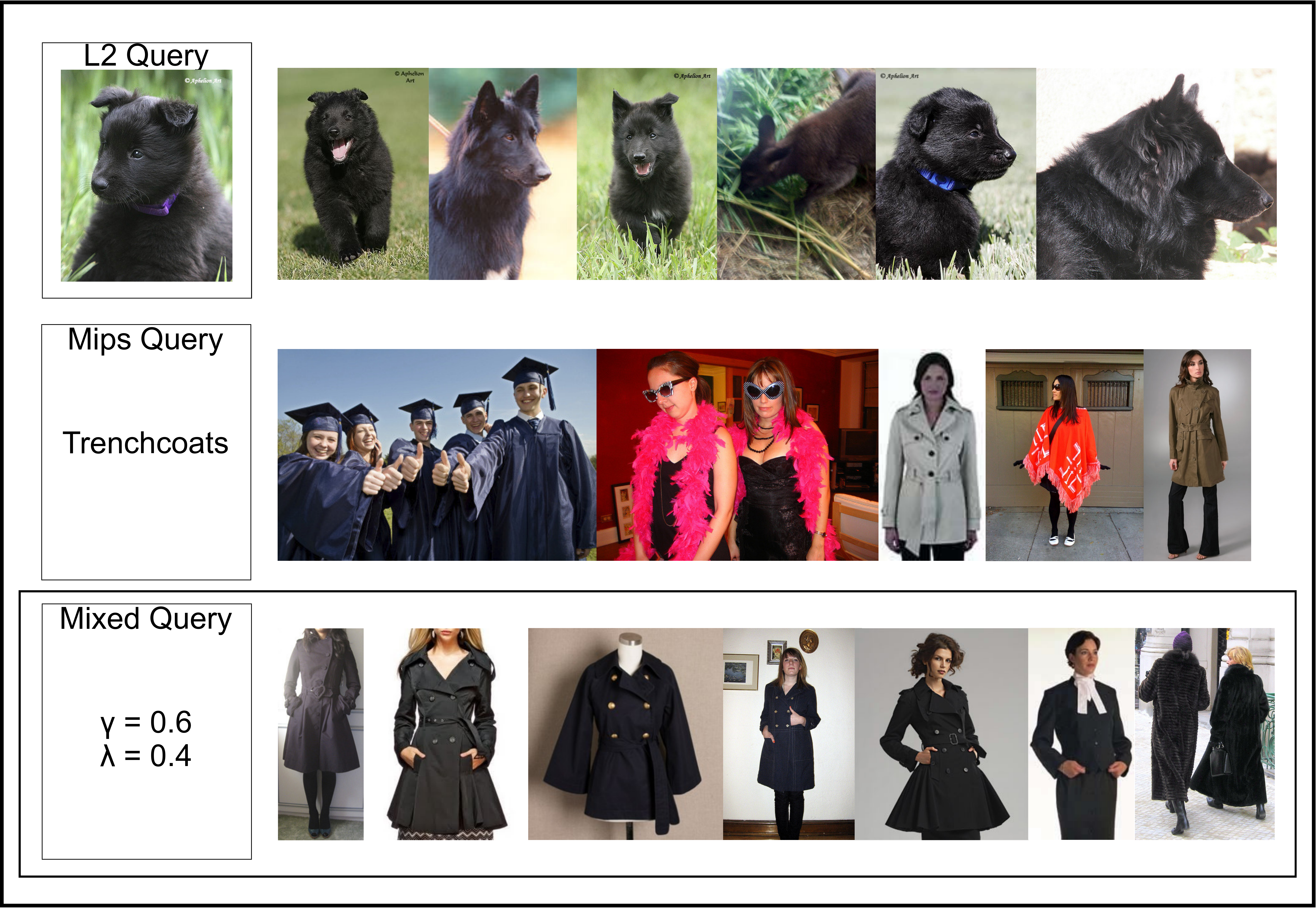}
    \label{fig:image_experiment.Trenchcoat}
  }
      \subfigure[Ice creams]{
        \includegraphics[width=\figsizeR]{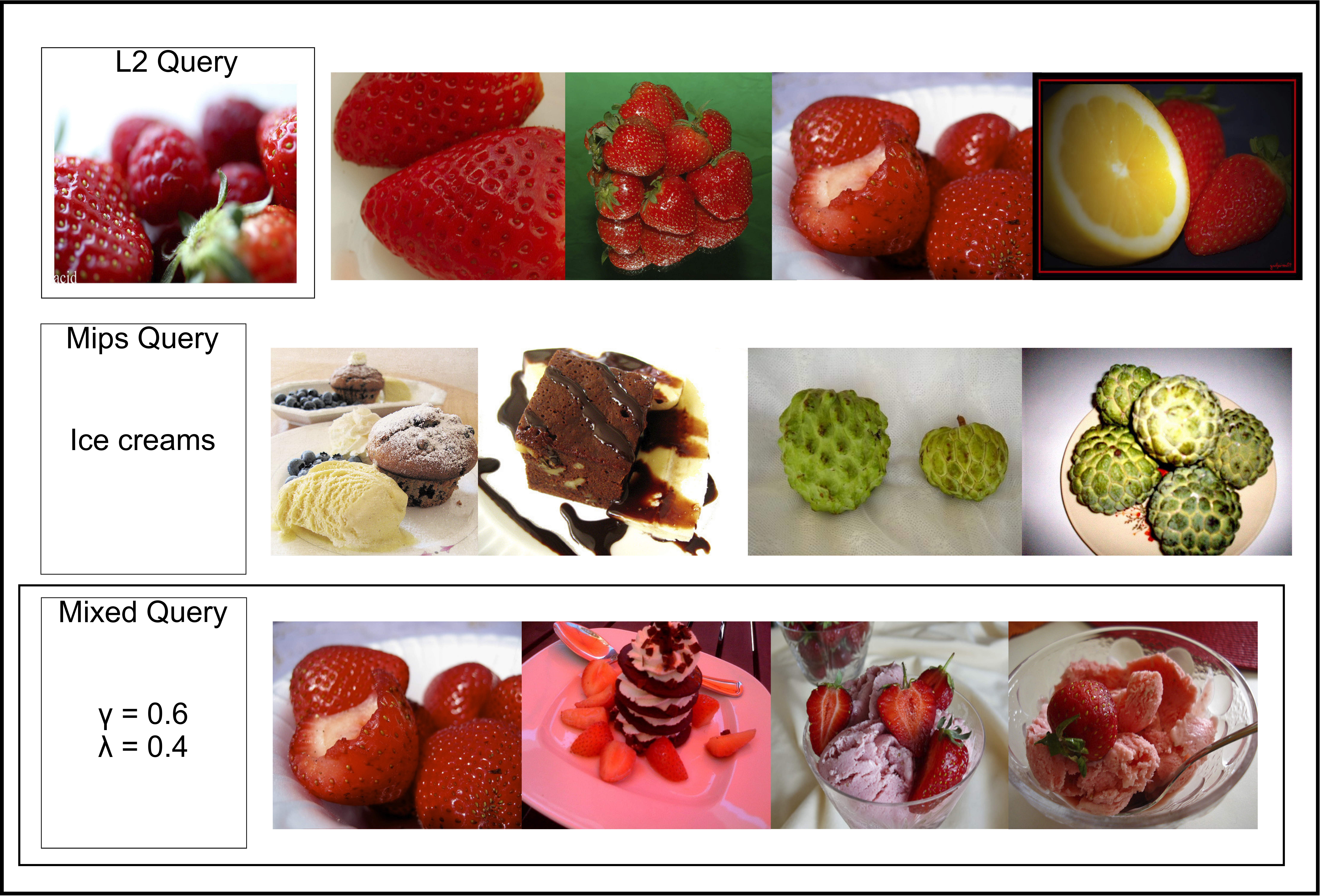}
    \label{fig:image_experiment.Icecream}
        }
  \caption{
  Image retrieval results with mixed queries.  In both of (a) and (b), the top row shows L2 query (left end) and the images retrieved (by ANNS with mp-LSH-CAT for $T=512$)
 according to the L2 dissimilarity ($\gamma^{(1)} = 1.0$ and $\lambda^{(2)} = 0.0$), 
 the second row shows MIPS query and the images retrieved according to the IP dissimilarity ($\gamma^{(1)} = 0.0$ and $\lambda^{(2)} = 1.0$),
 and the third row shows the images retrieved according to the mixed dissimilarity for $\gamma^{(1)} = 0.6$ and $\lambda^{(2)} = 0.4$.
  }
  \label{fig:image_experiment}
  \vspace{-2mm}
\end{figure*}

\subsection{Demonstration of Image Retrieval with Mixed Queries}
\label{sec:Experiment.Image}

Finally, we demonstrate the usefulness of our flexible mp-LSH in an image retrieval task on the ILSVRC2012 data set \cite{ILSVRC15}.
We computed a feature vector for each image by concatenating the 4096-dimensional fc7 activations of the trained VGG16 model \cite{Simonyan14c} with 120-dimensional color features\footnote{We computed histograms on the central crop of an image (covering 50\% of the area) for each rgb color channel with 8 and 32 bins. We normalized the histograms and concatenate them.}.
Since user preference vector is not available, we use classifier vectors, which are the weights associated with the respective ImageNet classes,
as MIPS queries
(the entries corresponding to the color features are set to zero). 
This simulates 
users who like a particular class of images.

We performed ANNS based on the MP dissimilarity by using our mp-LSH-CAT with $T= 512$ in the sample pool consisting of all $N \approx 1.2M$ images.
In Figure \ref{fig:image_experiment.Trenchcoat},
each of the three rows consists of the query at the left end, and the corresponding top-ranked images.
In the first row, the shown black dog image was used as the L2 query $\bfq^{(1)}$, and similar black dog images were retrieved
according to the L2 dissimilarity ($\gamma^{(1)} = 1.0$ and $\lambda^{(2)} = 0.0$).
In the second row, the VGG16 classifier vector for \emph{trench coats} was used as the MIPS query $\bfq^{(2)}$,
and images containing trench coats were retrieved according to the MIPS dissimilarity ($\gamma^{(1)} = 0.0$ and $\lambda^{(2)} = 1.0$).
In the third row, 
images containing black trench coats were retrieved according to 
the mixed dissimilarity for $\gamma^{(1)} = 0.6$ and $\lambda^{(2)} = 0.4$.
Figure \ref{fig:image_experiment.Icecream} shows another example with a strawberry L2 query and
the \emph{ice creams} MIPS query.
We see that, in both examples, mp-LSH-CAT handles the combined query well: it brings images that are close to the L2 query, and
relevant to the MIPS query.
Other examples can be found through our online demo.%
\footnote{\url{http://bbdcdemo.bbdc.tu-berlin.de/}}

\section{Conclusion}
\label{sec:Conclusion}

When querying huge amounts of data, it becomes mandatory to increase efficiency, i.e.,~even linear methods may be too computationally involved. Hashing, in particular locality sensitive hashing (LSH) has become
a highly efficient workhorse that can yield answers to queries in sublinear time, such as
L2-/cosine-distance nearest neighbor search (NNS) or maximum inner product search (MIPS).
While for typical applications the type of query has to be fixed beforehand, it is not uncommon to query with respect to several aspects in data, perhaps, even reweighting this dynamically at query time. 
Our paper contributes exactly herefore, namely by proposing three multiple purpose locality sensitive hashing (mp-LSH) methods 
\ifShortVersion
\else
which enable L2-/cosine-distance NNS, MIPS, and their weighted sums.%
\fi
\footnote{ 
Although a lot of hashing schemes for multi-modal data have been proposed \citep{Song13,Moran15,Xu13},
most of them are data-dependent, and do not offer adjustability of the importance weights
at query time.
}
A user can now indeed and efficiently change the importance of the weights at query time without recomputing the hash functions. 
\ifShortVersion
\else
Our paper has placed its focus on proving the feasibilty and efficiency of the mp-LSH methods, and introducing the very interesting cover tree concept
(which is less commonly applied in the machine learning world) 
for fast querying over the defined multi-metric space.
Finally we provide a demonstration on the usefulness of our novel technique.
\fi

Future studies will extend the possibilities of mp-LSH
for further including other types of dissimilarity measure, e.g., the distance from hyperplane \citep{Jain10},
and further applications with combined queires, e.g.,
retrieval with one complex multiple purpose query, say, a pareto-front for subsequent decision making. 
\ifShortVersion
\else
In addition we would like to analyze the interpretability of the nonlinear query mechanism in terms of salient features that have lead to the query result.
\fi

\appendices

%

\section{Derivation of Inner Product in Proof of Theorem~\ref{thrm:MPLSHVA}}
\label{sec:Proof.MPLSHVA}
The inner product between the augmented vectors $\widetilde{\bfq}$ and $\widetilde{\bfx}$, defined in Eq.\eqref{eq:MPLSHVA}, is given by
\begin{align*}
\widetilde{\bfq}^\top\widetilde{\bfx} 
&=
\textstyle
 \sum_{w=1}^W \sum_{g=1}^G
\left( (\gamma_g^{(w)} + \lambda_g^{(w)})\bfq_g^{(w)\T} \bfx_g
\right. \notag \\
& \hspace{20mm} \left.
\textstyle
 - \frac{1}{2}\sum_{g=1}^G \gamma_g^{(w)} \left(\|\bfq_g^{(w)}\|_2^2 + \|\bfx_g\|_2^2 \right) \right)\\
&=
\textstyle
 - \frac{1}{2}
\sum_{w=1}^W \sum_{g=1}^G
\Bigg( -2\lambda_g^{(w)} \bfq_g^{(w)\T} \bfx_g 
 \notag \\
& \hspace{15mm} 
\textstyle
+  \gamma_g^{(w)} \underbrace{\left((\|\bfq_g^{(w)}\|_2^2  + \|\bfx_g\|_2^2) - 2\bfq_g^{(w)\T} \bfx_g \right)}_{\|\bfq_g^{(w)} - \bfx_g \|_2^2} \Bigg) 
\notag\\
&= \textstyle
\pnorm{\bflambda}{1} - \frac{\mcL_{\mathrm{mp}}(\{\bfq^{(w)} \}, \bfx)}{2}.
\end{align*}

\section{Lemma: Inner Product Approximation}
\label{sec:IPApproximation}
For $\bfq, \bfx \in \mathbb{R}^L$ let
\[d_T(\bfq, \bfx) = \textstyle \frac{1}{T}\sum_{t=1}^{T}\left| \bfH(\bfq)_{t1} - \widetilde{\bfH}(\bfx)_{t1} \right|\]
with expectation
\[\textstyle
d(\bfq, \bfx) = \mathbb{E} d_T(\bfq, \bfx) = \mathbb{E}\left| \bfH(\bfq)_{11} - \widetilde{\bfH}(\bfx)_{11} \right|\]
and define
\[\textstyle
L(\bfq, \bfx) = 1 - \frac{\bfq^\T \bfx}{\pnorm{\bfq}{2}}.\]
\begin{lemma}
\label{lem:IPApproximation}
The following statements hold:

(a): It holds that \[\textstyle d(\bfq, \bfx) = 1 - \pnorm{\bfx}{2}(1 - \frac{2}{\pi}\sphericalangle(\bfq, \bfx))\]

(b): For $\CATError{x} = 0.2105\pnorm{\bfx}{2}$ it is
\begin{align}
|L(\bfq, \bfx) - d(\bfq, \bfx)| \leq \CATError{x}
\label{lem:IPApprox_b}
\end{align}

(c): Let $b(\bfq, \bfx) = 1 - \frac{2}{\pi}\frac{\bfq^\T\bfx}{\pnorm{\bfq}{2}}$, then for $L(\bfq, \bfx) \leq 1$ it is \[L(\bfq, \bfx) \leq d(\bfq, \bfx) \leq b(\bfq, \bfx) \leq 1\]
and for $L(\bfq, \bfx) \geq 1$ it is \[L(\bfq, \bfx) \geq d(\bfq, \bfx) \geq b(\bfq, \bfx) \geq 1\]

(d): It holds that \[|L(\bfq, \bfx) - d(\bfq, \bfx)| \leq \min\{(1 - \frac{2}{\pi})|L(\bfq, \bfx) - 1|, \CATError{x}\}\]
and for $s_{\bfx} = 0.58\pnorm{\bfx}{2}$, if $|L(\bfq, \bfx) - 1| \leq s_{\bfx}$, it is \[(1 - \frac{2}{\pi})|L(\bfq, \bfx) - 1| \leq \CATError{x}.\]
\end{lemma}
\noindent Proof (a):

Defining $p_{col} = 1 - \frac{1}{\pi}\sphericalangle(\bfq,\bfx)$ we have
\begin{align*}
 \mathbb{E}&\left| \bfH(\bfq)_{11} - \widetilde{\bfH}(\bfx)_{11} \right|\\
 &= \big(1 - \pnorm{\bfx}{2}\big)p_{col} + \big(1 + \pnorm{\bfx}{2}\big)\big(1 - p_{col}\big)\\
 &= 1 - \pnorm{\bfx}{2}\big(2p_{col}-1\big) = 1 - \pnorm{\bfx}{2}(1 - \frac{2}{\pi}\sphericalangle(\bfq, \bfx)).
\end{align*}

\noindent Proof (b):

\begin{align*}
|L(\bfq, \bfx) - d(\bfq, \bfx)| = \pnorm{\bfx}{2}| \frac{\bfq^\T \bfx}{\pnorm{\bfq}{2}\pnorm{\bfx}{2}} - 1 + \frac{2}{\pi}\sphericalangle(\bfq, \bfx)|\\
\leq \pnorm{\bfx}{2} \max_{z \in [-1,1]} | z - 1 + \frac{2}{\pi}\arccos(z)|.
\end{align*}
For $z^* = \sqrt{1 - \frac{4}{\pi^2}}$ we obtain the maximum
\[\CATError{x} = \pnorm{\bfx}{2}| z^* - 1 + \frac{2}{\pi}\arccos(z^*)| \approx 0.2105\pnorm{\bfx}{2}.\]

\noindent Proof (d):

The inequality follows from (b) and (c). Letting \[s_{\bfx} = \frac{\CATError{x}}{1 - \frac{2}{\pi}} \approx 0.58\pnorm{\bfx}{2},\]
the first bound is tighter than $\CATError{x}$, if $|L(\bfq, \bfx) - 1| \leq s_{\bfx}$.

\QED

Note that $d_T(\bfq, \bfx) \rightarrow d(\bfq, \bfx)$ as $T\rightarrow \infty$. Therefore all statements
are also valid, replacing $d(\bfq, \bfx)$ by $d_T(\bfq, \bfx)$ with $T$ large enough.

\section{Proof of Theorem~\ref{thrm:CATL2PlusIPApproximation}}
\label{sec:Proof.CATL2PlusIPApproximation}
For $\bfeta^{(w)} = \bfzero, \forall w$ we have
\[\mcL_{\mathrm{mp}}(\{\bfq^{(w)}\}, \bfx)
 = \sum_{w=1}^W   \sum_{g=1}^G \gamma_g^{(w)} \pnorm{\bfq_g^{(w)} - \bfx_g}{2}^2 - 2\lambda_g^{(w)} \bfq_g^{(w) \T} \bfx_g.\]
 Recall that $\overline{\bfq}_g^{\mathrm{L2+ip}} = \sum_{w=1}^W (\gamma_g^{(w)} + \lambda_g^{(w)}) \bfq_g^{(w)}$.
Therefore
\begin{align*}
&\frac{1}{T}\mcD_{\mathrm{CAT}} \Big(\bfH^{\mathrm{CAT-q}}(\{\bfq^{(w)}\}), \bfH^{\mathrm{CAT-x}}(\bfx) \Big)
 =  \sum_{g=1}^G\bigg(
 \frac{\overline{\gamma}_g}{2}\pnorm{\bfx_g}{2}^2 
 \\
 & \qquad \qquad
 +
\pnorm{\overline{\bfq}_g^{\mathrm{L2+ip}}}{2}\Big(1 + \pnorm{\bfx_g}{2}\big(1 - \frac{2}{T}\mcC_g(\overline{\bfq}^{\mathrm{L2+ip}}, \bfx)\big)\Big) \bigg).
\end{align*}
We use that
\begin{align*}
1 - \frac{2}{T}\mcC_g(\overline{\bfq}^{\mathrm{L2+ip}}, \bfx) &= -1 + \frac{1}{T} \sum_{t=1}^{T} \left| \bfH\left(\bfx\right)_{tg} - \bfH(\overline{\bfq}^{\mathrm{L2+ip}})_{tg} \right|\\
&\overset{\eqref{lem:IPApprox_b}}{\rightarrow} - \frac{\bfx_g^\T \overline{\bfq}_g^{\mathrm{L2+ip}}}{\pnorm{\bfx_g}{2}\pnorm{\overline{\bfq}_g^{\mathrm{L2+ip}}}{2}} + e_g,
\end{align*}
where $|e_g| \leq \CATError{1}$ such that
\begin{align*}
&\frac{1}{T}\mcD_{\mathrm{CAT}} \Big(\bfH^{\mathrm{CAT-q}}(\{\bfq^{(w)}\}), \bfH^{\mathrm{CAT-x}}(\bfx) \Big)
 = \sum_{g=1}^G\bigg(\\
 &\frac{\overline{\gamma}_g}{2}\pnorm{\bfx_g}{2}^2 +
\pnorm{\overline{\bfq}_g^{\mathrm{L2+ip}}}{2}\Big(1 - \frac{\bfx_g^\T \overline{\bfq}_g^{\mathrm{L2+ip}}}{\pnorm{\overline{\bfq}_g^{\mathrm{L2+ip}}}{2}}  + \pnorm{\bfx_g}{2}e_g\Big)\bigg)\\
&= \frac{1}{2}\sum_{g=1}^G \sum_{w=1}^W\left[ \gamma_g^{(w)} \pnorm{\bfq_g^{(w)} - \bfx_g}{2}^2 - 2\lambda_g^{(w)} \bfq_g^{(w) \T} \bfx_g\right]\\
&+ \underbrace{\sum_{g=1}^G \pnorm{\overline{\bfq}_g^{\mathrm{L2+ip}}}{2} - \frac{1}{2}\sum_{g=1}^G \sum_{w=1}^W \gamma_g^{(w)}\pnorm{\bfq_g^{(w)}}{2}^2}_{\text{const.}} 
\notag \\
& \qquad\qquad\qquad\qquad
+ \underbrace{\sum_{g=1}^G\pnorm{\overline{\bfq}_g^{\mathrm{L2+ip}}}{2}\pnorm{\bfx_g}{2}e_g}_{\text{error}}\\
&= \frac{1}{2}\mcL_{\mathrm{mp}}(\{\bfq^{(w)}\}, \bfx) - \pnorm{\bflambda}{1} + \text{ const } + \text{error}.
\end{align*}
We can bound the error-term by
\begin{align*}
|&\text{error}| \leq \max_{g \in \{1,\ldots,G\}} |e_g| \sum_{g=1}^G\pnorm{\overline{\bfq}_g^{\mathrm{L2+ip}}}{2}\pnorm{\bfx_g}{2}\\
&\leq \CATError{1}\pnormAdjustedBrackets{\left(\pnorm{\overline{\bfq}_g^{\mathrm{L2+ip}}}{2}\right)_g}{2}\pnorm{\bfx}{2} \leq \CATError{1}\pnormAdjustedBrackets{\left(\pnorm{\overline{\bfq}_g^{\mathrm{L2+ip}}}{2}\right)_g}{1}\\
&\leq \CATError{1}\sum_{g=1}^G\sum_{w=1}^W (\gamma_g^{(w)} + \lambda_g^{(w)})\pnorm{\bfq_g^{(w)}}{2} \leq \CATError{1}\left(\pnorm{\bflambda}{1} + \pnorm{\bfgamma}{1}\right).
\end{align*}
\QED

\section{Proof of Theorem~\ref{thrm:CATcosineApproximation}}
\label{sec:Proof.CATcosineApproximation}
For $\bfgamma^{(w)} = \bflambda^{(w)} = \bfzero, \forall w$ we have
\[\mcL_{\mathrm{mp}}(\{\bfq^{(w)}\}, \bfx)
 = -2\sum_{w=1}^W\sum_{g=1}^G \eta_g^{(w)} \frac{\bfq_g^{(w) \T} \bfx_g}{\pnorm{\bfq_g^{(w)}}{2}\pnorm{\bfx_g}{2}}.\]
Recall that $\overline{\bfq}_g^{\mathrm{cos}} = \sum_{w=1}^W \eta_g^{(w)} \frac{\bfq_g^{(w)}}{\pnorm{\bfq_g^{(w)}}{2}}$.
Therefore
\begin{align*}
\frac{1}{T}\mcD_{\mathrm{CAT}}& \Big(\bfH^{\mathrm{CAT-q}}(\{\bfq^{(w)}\}), \bfH^{\mathrm{CAT-x}}(\bfx) \Big)\\
&= \sum_{g=1}^G 2\pnorm{\overline{\bfq}_g^{\mathrm{cos}}}{2}\big(1 - \frac{1}{T}\mcC_g(\overline{\bfq}^{\mathrm{cos}}, \bfx)\big)\\
&\overset{\eqref{lem:IPApprox_b}}{\rightarrow} \sum_{g=1}^G \pnorm{\overline{\bfq}_g^{\mathrm{cos}}}{2}\big(1 - \frac{\bfx_g^\T \overline{\bfq}_g^{\mathrm{cos}}}{\pnorm{\bfx_g}{2}\pnorm{\overline{\bfq}_g^{\mathrm{cos}}}{2}} + e_g\big)\\
&= \underbrace{\sum_{g=1}^G \pnorm{\overline{\bfq}_g^{\mathrm{cos}}}{2}}_{\text{const.}} - \sum_{g=1}^G \frac{\bfx_g^\T \overline{\bfq}_g^{\mathrm{cos}}}{\pnorm{\bfx_g}{2}} + \underbrace{\sum_{g=1}^Ge_g\pnorm{\overline{\bfq}_g^{\mathrm{cos}}}{2}}_{\text{error}}\\
&= - \sum_{g=1}^G\sum_{w=1}^W \eta_g^{(w)} \frac{\bfx_g^\T \bfq_g^{(w)}}{\pnorm{\bfx_g}{2}\pnorm{\bfq_g^{(w)}}{2}}
+\text{const.} + \text{error}\\
&= \frac{1}{2}\mcL_{\mathrm{mp}}(\{\bfq^{(w)}\}, \bfx) - \pnorm{\bfeta}{1} + \text{const.} + \text{error}, 
\end{align*}
where
\begin{align*}
|\text{error}| &\leq \max_{g \in \{1,\ldots,G\}} |e_g| \sum_{g=1}^G\pnorm{\overline{\bfq}_g^{\mathrm{cos}}}{2}\\
&\leq \CATError{1}\sum_{g=1}^G\sum_{w=1}^W \eta_g^{(w)}\pnormAdjustedBrackets{\left.\bfq_g^{(w)} \middle/ \pnorm{\bfq_g^{(w)}}{2}\right.}{2}
= \CATError{1}\pnorm{\bfeta}{1}.
\end{align*}
 \QED
 
\section{Proof of Theorem~\ref{thrm:CATNNSGuarantee}}
\label{sec:Proof.CATNNSGuarantee}
Without loss of generality we prove the theorem for the plain MIPS case with
$G = 1$, $W = 1$ and $\lambda = 1$.
Then $\alpha = 1$ and the measure simplifies to
\begin{align*}
 \mcD_{\mathrm{CAT}} \Big(\bfH^{\mathrm{CAT-q}}(\{\bfq^{(w)}\}), \bfH^{\mathrm{CAT-x}}(\bfx) \Big) = Td_T(\bfq^{\mathrm{ip}},\bfx).
\end{align*}
For $\mcC_1(\bfq^{\mathrm{ip}}, \bfx)$ with $\mu = \mathbb{E}\mcC_1(\bfq^{\mathrm{ip}}, \bfx) = T(1 - \frac{1}{\pi}\sphericalangle(\bfx, \bfq^{\mathrm{ip}}))$ and $0 < \delta_1 < 1$, $\delta_2 > 0$ we use the following \emph{Chernoff}-bounds:
\begin{align}
 \mathbb{P}\big(\mcC_1(\bfq^{\mathrm{ip}}, \bfx) \leq (1-\delta_1)\mu\big) &\leq \exp\left\{-\frac{\mu}{2}\delta_1^2\right\}\label{ineq:chernoff1}\\
 \mathbb{P}\big(\mcC_1(\bfq^{\mathrm{ip}}, \bfx) \geq (1+\delta_2)\mu\big) &\leq \exp\left\{-\frac{\mu}{3}\min\{\delta_2, \delta_2^2\}\right\}\label{ineq:chernoff2}
\end{align}
The approximate nearest-neighbor problem with $r > 0$ and $c > 1$ is defined as follows:
If there exists an $\bfx^*$ with $\mcL_{\mathrm{ip}}(\bfq^{\mathrm{ip}},\bfx^*) \leq r$ then we return an $\widetilde{\bfx}$ with $\mcL_{\mathrm{ip}}(\bfq^{\mathrm{ip}}, \widetilde{\bfx}) < cr$.
For $cr > r + \CATError{1}$ we can set $T$ logarithmically dependent on the dataset size to solve the approximate nearest-neighbor problem for $\mcL_{\mathrm{ip}}$,
using $d_T$ with constant success probability:
For this we require a viable $t$ that fulfills 
\begin{align*}
 &\mcL_{\mathrm{ip}}(\bfq^{\mathrm{ip}},\bfx) > cr \Rightarrow d(\bfq^{\mathrm{ip}},\bfx) > t\text{ and}\\
 &\mcL_{\mathrm{ip}}(\bfq^{\mathrm{ip}},\bfx) \leq r \Rightarrow d(\bfq^{\mathrm{ip}},\bfx) <= t.
\end{align*}
Namely set $t = \frac{t_1 + t_2}{2}$, where
\begin{align*}
 &t_1  = \begin{cases}
          r + \CATError{1},&r\leq 1-s_{1}\\
          1 - \frac{2(1-r)}{\pi},& r\in(1-s_{1},1)\\
	  r,&r\geq 1
         \end{cases}\\
 \text{and }&t_2  = \begin{cases}
          cr,&cr\leq 1\\
          1 + \frac{2(cr-1)}{\pi},& cr\in(1,1+s_{1})\\
	  cr - \CATError{1},&cr\geq 1 + s_{1}
         \end{cases}.
\end{align*}
In any case it is $t_2 > t_1$:

First note that $t_1$ and $t_2$ are strictly monotone increasing in $r$ and $cr$, respectively.
It therefore suffices to show $\underline{t_2} \geq t_1$ for the lower bound $\underline{t_2}$ based on $\underline{cr} = r + \CATError{1}$.

(Case $r\leq 1-s_{1}$): It is $t_1 = r + \CATError{1}$ and $\underline{t_2} = \underline{cr}$, where
\[t_1 = r + \CATError{1} = \underline{cr} = \underline{t_2}\]

(Case $r\in(1-s_{1},1-\CATError{1}]$): It is $t_1 = 1 - \frac{2}{\pi}(1-r)$ and $\underline{t_2} = \underline{cr}$ such that
\begin{align*}
&t_1 = 1 - \frac{2}{\pi}(1-r) \leq r + \CATError{1} = \underline{cr} = \underline{t_2}\\
\Leftrightarrow &(1 - \frac{2}{\pi})(1-r) \leq \CATError{1} \Leftrightarrow (1-r) \leq s_1 \Leftrightarrow r \geq 1 - s_1
\end{align*}

(Case $r\in(1-\CATError{1},1]$): It is $t_1 = 1 - \frac{2}{\pi}(1-r)$ and $\underline{t_2} = 1 + \frac{2}{\pi}(\underline{cr} - 1)$ with $\underline{cr} > 1$ such that
\begin{align*}
t_1 = 1 - \frac{2}{\pi}(1-r) \leq 1 \leq 1 + \frac{2}{\pi}(\underline{cr} - 1) = \underline{t_2}
\end{align*}

(Case $r\in(1,1+s_1-\CATError{1}]$): It is $t_1 = r$ and $\underline{t_2} = 1 + \frac{2}{\pi}(\underline{cr} - 1)$ such that
\begin{align*}
&t_1 = r \leq 1 + \frac{2}{\pi}(r + \CATError{1} - 1) = 1 + \frac{2}{\pi}(\underline{cr} - 1) = \underline{t_2}\\
\Leftrightarrow &(1-\frac{2}{\pi})r \leq (1-\frac{2}{\pi}) - (1-\frac{2}{\pi})\CATError{1} + \CATError{1}\\
\Leftrightarrow &r \leq 1 + s_1 - \CATError{1}
\end{align*}

(Case $r > 1+s_1-\CATError{1}$): It is $t_1 = r$ and $\underline{t_2} = \underline{cr} - \CATError{1}$, where
\[t_1 = r = \underline{cr} - \CATError{1} = \underline{t_2}\]\QED

Now, define
\[\delta = \left|\frac{t - d(\bfq^{\mathrm{ip}},\bfx)}{1+\pnorm{\bfx}{2}-d(\bfq^{\mathrm{ip}},\bfx)}\right| = \left|T\frac{t - d(\bfq^{\mathrm{ip}},\bfx)}{2\pnorm{\bfx}{2}\mu}\right|.\]

For $\mcL_{\mathrm{ip}}(\bfq^{\mathrm{ip}},\bfx) \leq r$ we can lower bound the probability of $d_T(\bfq^{\mathrm{ip}},\bfx)$ not exceeding the specified threshold:
\begin{align*}
  \mathbb{P}\big(&d_T(\bfq^{\mathrm{ip}},\bfx) \leq t  \big) = \mathbb{P}\big( \mcC(\bfq^{\mathrm{ip}}, \bfx) \geq (1-\delta)\mu\big)\\
  &= 1 - \mathbb{P}\big( \mcC(\bfq^{\mathrm{ip}}, \bfx) \leq (1-\delta)\mu\big) \overset{\eqref{ineq:chernoff1}}{\geq} 1 - \exp\left\{-\frac{\mu}{2}\delta^2\right\}.
\end{align*}
We can show $d(\bfq^{\mathrm{ip}},\bfx) \leq t_1$, using Lemma~\ref{lem:IPApproximation}, (c) and (d):

(Case $r\leq 1-s_{1}$):
\[d(\bfq^{\mathrm{ip}},\bfx) - \mcL_{\mathrm{ip}}(\bfq^{\mathrm{ip}},\bfx) \leq \CATError{1} \Rightarrow d(\bfq^{\mathrm{ip}},\bfx) \leq r + \CATError{1}\]

(Case $r \in (1-s_{1},1)$):
\begin{align*}
 &d(\bfq^{\mathrm{ip}},\bfx) - \mcL_{\mathrm{ip}}(\bfq^{\mathrm{ip}},\bfx) \leq (1-\frac{2}{\pi})(1 - \mcL_{\mathrm{ip}}(\bfq^{\mathrm{ip}},\bfx))\\
 \Rightarrow\,&d(\bfq^{\mathrm{ip}},\bfx) \leq 1 - \frac{2}{\pi}(1 - \mcL_{\mathrm{ip}}(\bfq^{\mathrm{ip}},\bfx)) \leq 1 - \frac{2}{\pi}(1 - r) = t_1
\end{align*}

(Case $r \geq 1$): For $\mcL_{\mathrm{ip}}(\bfq^{\mathrm{ip}},\bfx) \leq 1$ it is $d(\bfq^{\mathrm{ip}},\bfx) \leq 1$.
Else $d(\bfq^{\mathrm{ip}},\bfx) \leq \mcL_{\mathrm{ip}}(\bfq^{\mathrm{ip}},\bfx)$ such that
\[d(\bfq^{\mathrm{ip}},\bfx) \leq \max\{1, \mcL_{\mathrm{ip}}(\bfq^{\mathrm{ip}},\bfx)\} \leq r = t_1\]
Thus we can bound
\[\delta \overset{d(\bfq^{\mathrm{ip}},\bfx) \leq t_1 < t}{\geq} \frac{T(t - t_1)}{2\pnorm{\bfx}{2}\mu} \overset{\pnorm{\bfx}{2} \leq 1}{\geq} \frac{T(t - t_2)}{2\mu} = \frac{T(t_2-t_1)}{4\mu}\]
and
\[\delta^2\mu \geq \frac{T^2(t_2-t_1)^2}{16\mu} \overset{\mu\leq T}{\geq} \frac{T(t_2-t_1)^2}{16},\]
such that
\[\mathbb{P}\big( d_T(\bfq^{\mathrm{ip}},\bfx) \leq t  \big) \geq 1 - \exp\left\{-\frac{(t_2-t_1)^2}{32}T\right\}.\]

For $\mcL_{\mathrm{ip}}(\bfq^{\mathrm{ip}},\bfx) > cr$ we can upper bound the probability of $d_T(\bfq^{\mathrm{ip}},\bfx)$ dropping below the specified threshold:
\begin{align*}
  \mathbb{P}\big( d_T(\bfq^{\mathrm{ip}},\bfx) \leq t  \big) = \mathbb{P}\big( \mcC(\bfq^{\mathrm{ip}}, \bfx) \geq (1+\delta)\mu\big)\\
 \overset{\eqref{ineq:chernoff2}}{\leq} \exp\left\{-\frac{\mu}{3}\min\{\delta, \delta^2\}\right\}.
\end{align*}
We can show $d(\bfq^{\mathrm{ip}},\bfx) \geq t_2$, using Lemma~\ref{lem:IPApproximation}, (c) and (d):

(Case $cr\leq 1$): For $\mcL_{\mathrm{ip}}(\bfq^{\mathrm{ip}},\bfx) \geq 1$ it is $d(\bfq^{\mathrm{ip}},\bfx) \geq 1$.
Else $d(\bfq^{\mathrm{ip}},\bfx) \geq \mcL_{\mathrm{ip}}(\bfq^{\mathrm{ip}},\bfx)$ such that
\[d(\bfq^{\mathrm{ip}},\bfx) \geq \min\{1, \mcL_{\mathrm{ip}}(\bfq^{\mathrm{ip}},\bfx)\} \geq cr = t_2\]

(Case $cr\in (1, 1+s_1)$):
\begin{align*}
 &\mcL_{\mathrm{ip}}(\bfq^{\mathrm{ip}},\bfx) - d(\bfq^{\mathrm{ip}},\bfx) \leq (1-\frac{2}{\pi})(\mcL_{\mathrm{ip}}(\bfq^{\mathrm{ip}},\bfx) - 1)\\
 \Rightarrow\,&d(\bfq^{\mathrm{ip}},\bfx) \geq 1 + \frac{2}{\pi}(\mcL_{\mathrm{ip}}(\bfq^{\mathrm{ip}},\bfx) - 1) \geq 1 - \frac{2}{\pi}(cr - 1) = t_2
\end{align*}

(Case $cr \geq 1+s_1$):
\[\mcL_{\mathrm{ip}}(\bfq^{\mathrm{ip}},\bfx) - d(\bfq^{\mathrm{ip}},\bfx) \leq \CATError{1} \Rightarrow d(\bfq^{\mathrm{ip}},\bfx) \geq cr - \CATError{1} = t_2\]
Thus we can bound
\[\delta \overset{d(\bfq^{\mathrm{ip}},\bfx) \geq t_2 > t}{\geq} \frac{T(t_2 - t)}{2\pnorm{\bfx}{2}\mu} \overset{\pnorm{\bfx}{2} \leq 1}{\geq} \frac{T(t_2 - t)}{2\mu} = \frac{T(t_2 - t_1)}{4\mu},\]
such that
\begin{align*}
 \mathbb{P}\big( d_T(\bfq^{\mathrm{ip}}&,\bfx) \leq t  \big) \leq \textstyle  \exp\left\{-\min\left\{\frac{T(t_2 - t_1)}{12}, \frac{T^2(t_2 - t_1)^2}{48\mu}\right\}\right\}\\
 \overset{\mu\leq T}{\leq} & \textstyle \exp\left\{-\min\left\{\frac{T(t_2 - t_1)}{12}, \frac{T(t_2 - t_1)^2}{48}\right\}\right\}\\
 = & \textstyle \exp\left\{-\frac{T}{3}\min\left\{\frac{t_2 - t_1}{4}, \left(\frac{t_2 - t_1}{4}\right)^2\right\}\right\}\\
 \overset{\frac{t_2 - t_1}{4} < 1}{=} & \textstyle \exp\left\{-\frac{T}{3}\left(\frac{t_2 - t_1}{4}\right)^2\right\} = \exp\left\{-\frac{(t_2 - t_1)^2}{48}T\right\}.
\end{align*}

Now, define the events
\begin{align}
 &E_1(\bfq^{\mathrm{ip}},\bfx):\text{ either } \mcL_{\mathrm{ip}}(\bfq^{\mathrm{ip}}, \bfx) > r\text{ or } d_T(\bfq^{\mathrm{ip}}, \bfx) \leq t\\
 &E_2(\bfq^{\mathrm{ip}}):\forall \bfx\in X:\mcL_{\mathrm{ip}}(\bfq^{\mathrm{ip}}, \bfx) > cr \Rightarrow d_T(\bfq^{\mathrm{ip}}, \bfx) > t
\end{align}
Assume that there exists $\bfx^*$ with $\mcL_{\mathrm{ip}}(\bfq^{\mathrm{ip}}, \bfx^*) \leq r$. Then the algorithm is successful if both, $E_1(\bfq^{\mathrm{ip}}, \bfx^*)$ and $E_2(\bfq^{\mathrm{ip}})$
hold simultaneously. Let $T \geq \frac{48}{(t_2-t_1)^2}\log(\frac{n}{\varepsilon})$. It is
\begin{align*}
 \mathbb{P}\big(E_2(\bfq^{\mathrm{ip}})\big) & \!=\! 1 \!- \! \mathbb{P}\big(\exists \bfx\in X: \mcL_{\mathrm{ip}}(\bfq^{\mathrm{ip}}, \bfx) \!>\! cr, d_T(\bfq^{\mathrm{ip}}, \bfx^*)\! \leq \! t\big)\\
 &\geq 1 - \sum_{\bfx \in X} \mathbb{P}\big(\mcL_{\mathrm{ip}}(\bfq^{\mathrm{ip}}, \bfx) > cr, d_T(\bfq^{\mathrm{ip}}, \bfx) \leq t\big)\\
 &\geq \textstyle  1 - n\exp\left\{-\frac{(t_2-t_1)^2}{48}T\right\} \geq 1 - \varepsilon.
\end{align*}
Also it holds
\begin{align*}
 \mathbb{P}\big(E_1(\bfq^{\mathrm{ip}}, \bfx^*)\big) \geq   1 - \left(\frac{\varepsilon}{n}\right)^\frac{3}{2}.
\end{align*}
Therefore the probability of the algorithm to perform approximate nearest neighbor search correctly is larger than
\begin{align*}
 &\mathbb{P}\big(E_2(\bfq^{\mathrm{ip}}),  E_1(\bfq^{\mathrm{ip}}, \bfx^*)\big)\\
 &\geq 1 - \mathbb{P}\big(\neg E_2(\bfq^{\mathrm{ip}})\big) - \mathbb{P}\big(\neg E_1(\bfq^{\mathrm{ip}}, \bfx^*)\big)
 \geq 1 - \varepsilon - \left(\frac{\varepsilon}{n}\right)^\frac{3}{2}.
\end{align*}

\section{Details of Cover Tree}
\label{sec:CoverTreeDetail}

Here, we detail how to selectively explore the hash buckets with 
the code dissimilarity measure 
in non-increasing order.
%
%
%
The difficulty is in that 
the dissimilarity 
$\mcD$ is a \emph{linear combination} of metrics, where the weights 
are selected at query time. 
Such a metric is referred to as a \emph{dynamic metric function} or a \emph{multi-metric} \cite{Bustos12}.
We use a tree data structure, called the 
\emph{cover tree} \cite{Beygelzimer06},
 to index the metric space.

We begin the description of the cover tree by introducing the \emph{expansion constant} and the \emph{base of the expansion 
	constant}.

\textbf{Expansion Constant ($\kappa$)} \cite{Heinonen01}:
is defined as the smallest value $\kappa \geq \psi $ such that 
every ball in the dataset $\mcX$ can be covered by $\kappa$ balls in $\mcX$ of
radius equal $1/\psi$. Here, $\psi$ is the \emph{base of the expansion constant}.

%

\textbf{Data Structure:} Given a set of data points $\mcX$, the 
cover tree $\mcT$ is a leveled tree where each level is associated with an integer
label $\ctlevel$, which decreases as the tree is descended. 
For ease of explanation, let $B_{\psi^\ctlevel}(\bfx)$ denote a \emph{closed ball} centered at 
point $\bfx$ with radius $\psi^\ctlevel$, i.e., $B_{\psi^\ctlevel}(\bfx) = \{ p \in \mcX: \mcD(p,\bfx) \leq \psi^\ctlevel \}$. At every level $\ctlevel$ of $\mcT$ (except the root), we create a \emph{union of possibly overlapping closed balls} with radius $\psi^\ctlevel$ that \emph{cover} (or contain) all the data points $\mcX$. The centers of this covering set of balls are stored in \emph{nodes} at level $\ctlevel$ of $\mcT$. Let $\mcC_\ctlevel$ denote the set of nodes at level $\ctlevel$. 
The cover tree $\mcT$ obeys the following three invariants at all levels:

\begin{enumerate}
	\item (\textbf{Nesting}) $\mcC_\ctlevel \subset \mcC_{\ctlevel-1}$. 
	Once a point $\bfx \in \mcX$ is in a node in $\mcC_\ctlevel$, then it also
	appears in all its successor nodes.
	
	\item (\textbf{Covering}) For every $\bfx' \in \mcC_{\ctlevel-1}$, there exists a $\bfx \in \mcC_{\ctlevel}$
	where $\bfx'$ lies inside $B_{\psi^\ctlevel}(\bfx)$, and exactly one such 
	$\bfx$ is a parent of $\bfx'$.
	
	
	\item (\textbf{Separation}) For all $\bfx_1,\bfx_2 \in \mcC_{\ctlevel}$, 
	$\bfx_1$ lies outside $B_{\psi^\ctlevel}(\bfx_2)$ and 
	$\bfx_2$ lies outside $B_{\psi^\ctlevel}(\bfx_1)$.
	
	
\end{enumerate}
This structure has a space bound of $O(N)$, where $N$ is the number of samples.

\textbf{Construction:} We use the \emph{batch} construction method
\citep{Beygelzimer06},
where
the cover tree $\mcT$ is built in a \emph{top-down} fashion. 
Initially, we pick a data point $\bfx^{(0)}$ and an integer $s$, such that the closed ball $B_{\psi^{s}} (\bfx^{(0)})$ is the tightest fit that covers the entire dataset $\mcX$.

This point $\bfx^{(0)}$ is placed in a single node, called the \emph{root} of the tree $\mcT$. We denote the root node as $\mcC_{\ctlevel}$ (where $\ctlevel = s$).
In order to generate the set $\mcC_{\ctlevel-1}$ of the child nodes for $\mcC_\ctlevel$,
we greedily pick a set of points (including point $\bfx^{(0)}$ from $\mcC_\ctlevel$ to satisfy the 
\emph{Nesting} invariant) and generate closed balls of radius $\psi^{\ctlevel-1}$ centered on them,
in such a way that: (a) all center points lie inside $B_{\psi^\ctlevel}(\bfx^{(0)})$
(\emph{Covering} invariant), 
(b) no center point intersects with other balls of radius $\psi^{\ctlevel-1}$ at  
level $\ctlevel-1$ (\emph{Separation} invariant),
and (c) the union of these closed balls covers the entire dataset $\mcX$.
These chosen center points form the set of nodes $\mcC_{\ctlevel-1}$.
Child nodes are \emph{recursively} generated from each node in $\mcC_{\ctlevel-1}$,
until each data point in $\mcX$ is the center of a closed ball 
and resides in a leaf node of $\mcT$.

Note that, while we construct our cover tree, we 
use our distance function $\mcD$ with all the weights set to $1.0$, which
upper bounds all subsequent distance metrics that depend on the queries.
The construction time complexity is $O(\kappa^{12} N \ln N)$.

To achieve a more compact cover tree, we store only element identification numbers (IDs) in the cover tree, and not the original vectors. Furthermore, we store the hash bits using \emph{compressed representation bit-sets} that reduce the storage size compared to a naive implementation down to 
$T$ bits. 
For mp-LSH-CA with $G=1$, each element in the cover tree contains 
$T$ bits and $2$ integers. 
For example, indexing a $128$ dimensional vector naively requires 1032 bytes, but indexing the fully augmented one requires only 24 bytes, yielding a $97.7\%$ memory saving.%
\footnote{We assume 4 bytes per integer and 8 bytes per double here.}

\ifOmitCovertreeAlgorithm

%
%
%
%

\textbf{Querying}:
The nearest neighbor search with a cover tree is performed as follows.
The search for the nearest neighbor begins at the root of the cover tree and 
descends level-wise.
On each descent, we build a candidate set $\mcC$, which holds 
all the child nodes (center points of our closed balls).
We then \emph{prune} away centers (nodes) in $\mcC$ that cannot possibly lead to a nearest neighbor to the query point $\bfq$, if we descended down them.

The pruning mechanism is predicated on a proven result in~\cite{Beygelzimer06} 
which states that for any point $\bfx \in \mcC_{\ctlevel-1}$, the distance between $\bfx$
and any descendant $\bfx'$ is upper bounded by $\psi^\ctlevel$. 
Therefore, 
any center point whose distance from $\bfq$ exceeds $\min_{\bfx' \in \mcC}\mcD(\bfq, \bfx') + \psi^\ctlevel$ cannot possibly have a descendant that can replace the current closest center point to $\bfq$ and hence can safely be pruned. 
We add an additional check to speedup the search by not always 
descending to the leaf node. The time complexity of querying the cover tree is $O(\kappa^{12} \ln N)$. 

\fi

\textbf{Effect of multi-metric distance while querying}: 
It is important to note that minimizing overlap between the
closed balls on higher levels (i.e., closer to the root) of the cover tree 
can allow us to effectively prune a very large portion of the search space and compute the nearest neighbor faster.

Recall that the cover tree is constructed by setting our distance function $\mcD$ with all the weights set to $1.0$. During querying, we allow $\mcD$ to be a linear combination of metrics, where the weights lie in the range $[0,1]$, which means that the distance metric $\mcD$ used during querying always \emph{under-estimates} the distances and reports lower distances.
During querying, the cover tree's structure is still intact and all the invariant 
properties satisfied. 
The main difference occurs 
\ifOmitCovertreeAlgorithm
in computation of $\min_{\bfx' \in \mcC}\mcD(\bfq, \bfx')$, which is the shortest distance from a center point to the query $\bfq$ (using the new distance metric).
\fi
Interestingly, this new distance gets even smaller, thus reducing our search radius 
(i.e., $\min_{\bfx' \in \mcC}\mcD(\bfq, \bfx') + \psi^\ctlevel$) centered at $\bfq$, which in 
turn implies that at every level we manage to prune more center points, as the overlap
between the closed balls also is reduced.

\textbf{Streaming}: The cover tree lends itself naturally to the setting where
nearest neighbor computations have to be performed on a stream of data points. 
This is because the cover tree allows dynamic insertion and deletion of points. 
The time complexity for both these operations is $O(\kappa^{6} \ln N)$, which 
is faster than querying.

\textbf{Parameter choice}: 
In our implementation for experiment,
we set the \emph{base of expansion constant} to $\psi = 1.2$,
which we empirically found to work best on the texmex dataset. 

\ifOmitAdditionalTexMexResult
\else

\section{Additional Information on Computation Time Evaluation}
\label{sec:AdditionalExperimentalResultComputationTime}

\ifShortVersion

Tables~\ref{tb:ResultCCApp}--\ref{tb:ResultCCSameMemoryApp} summarize
recall@$k$, query time, cover tree construction time, and required memory storage
(recall@$k$ and query time are the same as in Tables~\ref{tb:ResultCC}--\ref{tb:ResultCCSameMemory} in the main text).
In our implementation,
the gap in the memory requirement between mp-LSH-CC and mp-LSH-CA tends to be larger than the theory expects.
This is because efficient coding for mp-LSH-CA is easier than for mp-LSH-CC.
Query time by mp-LSH-CC is increased when the number $T$ of hash bits is decreased (compare Tables~\ref{tb:ResultCCApp} and \ref{tb:ResultCCSameMemoryApp}).
This happened because $T$ is too small to distinguish samples, which consequently increased the number of distance evaluations in cover tree querying.

\begin{table*}[t]
\centering
\caption{ANNS Results with mp-LSH-CC with $T=128$ ($N = 10^8$).
}
\label{tb:ResultCCApp}
\begin{tabular}{@{}rrrrrrrrr@{}}
\toprule
        & \multicolumn{3}{l}{Recall@$k$} & \multicolumn{3}{l}{Query time (msec)} & \multirow{2}{*}{\begin{tabular}[c]{@{}l@{}}Cover Tree\\ Construction (sec)\end{tabular}} & \multirow{2}{*}{\begin{tabular}[c]{@{}l@{}}Storage Requirement\\ per sample (bytes)\end{tabular}} \\
        & 1         & 5         & 10       & 1           & 5           & 10          &          &  \\ \midrule

L2 & 0.98 & 1.00 & 1.00 & 119.38 & 132.08 & 146.06 & 18638 & 632 \\

MIPS & 0.74 & 0.80 & 0.82 & 205.95 & 207.54 & 207.86 & 18638 & 632 \\

L2+MIPS & 0.29 & 0.59 & 0.62 & 186.94 & 190.87 & 191.74 & 18638 & 632 \\\bottomrule
\end{tabular}

\centering
\caption{ANNS Results with mp-LSH-CA with $T=128$ ($N = 10^8$).
}
\label{tb:ResultCAApp}
\begin{tabular}{@{}rrrrrrrrr@{}}
\toprule
        & \multicolumn{3}{l}{Recall@$k$} & \multicolumn{3}{l}{Query time (msec)} & \multirow{2}{*}{\begin{tabular}[c]{@{}l@{}}Cover Tree\\ Construction (sec)\end{tabular}} & \multirow{2}{*}{\begin{tabular}[c]{@{}l@{}}Storage Requirement\\ per sample (bytes)\end{tabular}} \\
        & 1         & 5         & 10       & 1           & 5           & 10          &          &  \\ \midrule

L2 & 0.58 & 0.94 & 1.00 & 0.04 & 0.10 & 0.07 & 11585 & 104 \\

MIPS & 0.56 & 0.59 & 0.68 & 8.32 & 4.96 & 5.67 & 11585 & 104 \\

L2+MIPS & 0.27 & 0.77 & 0.88 & 111.35 & 130.74 & 146.87 & 11585 & 104 \\\bottomrule\end{tabular}

\centering
\caption{ANNS Results with mp-LSH-CC with $T=43$ ($N = 10^8$).
}
\label{tb:ResultCCSameMemoryApp}
\begin{tabular}{@{}rrrrrrrrr@{}}
\toprule
        & \multicolumn{3}{l}{Recall@$k$} & \multicolumn{3}{l}{Query time (msec)} & \multirow{2}{*}{\begin{tabular}[c]{@{}l@{}}Cover Tree\\ Construction (sec)\end{tabular}} & \multirow{2}{*}{\begin{tabular}[c]{@{}l@{}}Storage Requirement\\ per sample (bytes)\end{tabular}} \\
        & 1         & 5         & 10       & 1           & 5           & 10          &          &  \\ \midrule

L2 & 0.61 & 0.81 & 0.87 & 143.65 & 162.83 & 163.83 &  4642 & 296 \\

MIPS & 0.06 & 0.12 & 0.18 & 235.06 & 238.66 & 242.34 & 4642 & 296 \\

L2+MIPS & 0.15 & 0.25 & 0.30 & 226.25 & 230.40 & 232.44 & 4642 & 296 \\\bottomrule
\end{tabular}
\end{table*}

\fi

Tables~\ref{perftable} and \ref{perftable2} summarize the cover tree construction time
and the query time with mp-LSH-CA for $T=128$, compared with the brute force search time, for different number $N$ of samples.
We can observe the sub-linear nature of the query time.

Table~\ref{covertreealone} shows the computation time of L2-NNS with the cover tree applied to the metric in the original space (without using any LSH coding).
For cover tree without LSH,
we need to make the IP dissimilarity, the third term in the MP dissimilarity \eqref{eq:Objective}, to be a metric
by using the augmentation, Eqs.\eqref{eq:SimpleLSHQ} and \eqref{eq:SimpleLSHX}, used for simple-LSH.
Since
\[
\|\widetilde{\bfq} - \widetilde{\bfx}\|_2^2
= - 2 \bfq^\T \bfx + \mathrm{const},
\]
the following multi-metric keep the same ordering as the MP dissimilarity:
 \begin{align}
 &\mcL_{\mathrm{mp}}(\{\bfq^{(w)}\}, \bfx)
 = \textstyle \sum_{w=1}^W   \sum_{g=1}^G \bigg\{  \gamma_g^{(w)} \|\bfq_g^{(w)} - \bfx_g\|_2^2
 \notag \\
 & \hspace{-2mm}
 \textstyle
  + 2 \eta_g^{(w)} \left(1 -  \frac{\bfq_g^{(w) \T} \bfx_g}{\|\bfq_g^{(w)} \|_2 \|\bfx_g\|_2}\right)
 +  \lambda_g^{(w)} \|\widetilde{\bfq}_g^{(w)} - \widetilde{\bfx}_g\|_2^2 \bigg\}.
\label{eq:ObjectiveWithoutLSH}
 \normalsize
 \end{align}
This approach is guaranteed to provide the exact NNS solution, but comparing Tables~\ref{perftable} and \ref{covertreealone} implies 
that the query time is $500 \sim 1000$ times slower than our approach with mp-LSH-CA.
This is because the distance in the original space needs to be evaluated many times when the algorithm goes down to the leaves of the cover tree.
In addition, this approach requires significantly more memory storage (2144 bytes/sample for cover tree alone vs. 104 bytes/sample for mp-LSH-CA with $T=128$).

%

\begin{table*}[t]
	\centering
	\caption{Performance Chart for L2NN search with mp-LSH-CAT, T=1024}
	\label{perftable_L2}
	\begin{tabular}{@{}r|r|rrrrr|r@{}}
		\toprule
		Datasize                      & Index Build Time & \multicolumn{5}{c|}{Query Time for knn (ms)} & Brute Force time \\
		& ms               & 1       & 2       & 3      & 5      & 10     & ms                      \\ \midrule
		100 &                      29 &     0.03&     0.00&     0.00&     0.01&     0.00&     0.84 \\
		1,000 &                      69 &     0.07&     0.01&     0.02&     0.03&     0.03&     1.17                    \\
		10,000 &                     717 &     3.56&     3.00&     2.99&     3.15&     3.34&     6.09                    \\
		100,000 & 35146 &    53.37&    54.37&   55.33&   56.24&    57.38&    58.39                   \\
1,000,000& 2350340 &   461.50&   477.46&   484.02&   491.64&   501.06 & 667.02                  \\
		
		10,000,000 & 41957790& 583.85 & 601.83 & 609.15 & 617.02 & 626.02 & 19982.22                \\ \bottomrule
\end{tabular}
	\centering
	\caption{Performance Chart for MIPS with mp-LSH-CAT, T=1024}
	\label{perftable_MIPS}
	\begin{tabular}{@{}r|r|rrrrr|r@{}}
		\toprule
		Datasize                      & Index Build Time & \multicolumn{5}{c|}{Query Time for knn (ms)} & Brute Force time \\
		& ms               & 1       & 2       & 3      & 5      & 10     & ms                      \\ \midrule
		100 &                      29 &0.00&     0.00&     0.00&     0.00&     0.00&     7.75 \\
		1,000 &                      69 &     0.00&     0.00&     0.00&     0.00&     0.00&     8.15                    \\
		10,000 &                     717 &     3.19&     2.73&     2.80&     2.96&     3.07&     14.08                    \\
		100,000 &                   35146 &    40.30&    42.27&    43.08&    43.81&    44.77&    84.60                   \\
1,000,000& 2350340 &   434.19&   455.61&   462.87&   469.71&   479.34&   768.98                  \\
		
		10,000,000 & 41957790& 593.11& 619.22& 627.93& 635.72& 645.14 & 7748.15                \\ \bottomrule
\end{tabular}
	\centering
	\caption{Performance Chart for MIPS+L2 with mp-LSH-CAT, T=1024}
	\label{perftable_MIPS_L2}
	\begin{tabular}{@{}r|r|rrrrr|r@{}}
		\toprule
		Datasize                      & Index Build Time & \multicolumn{5}{c|}{Query Time for knn (ms)} & Brute Force time \\
		& ms               & 1       & 2       & 3      & 5      & 10     & ms                      \\ \midrule
		100 &                      29 &0.00&     0.00&     0.00&     0.00&     0.00&     0.10 \\
		1,000 &                      69 &     0.00&     0.00&     0.00&     0.00&     0.00&     0.71                    \\
		10,000 &                     717 &     2.94&     2.47&     2.55&     2.66&     2.77&     10.08                    \\
		100,000 &                   35146 &    43.81&    44.46&    45.25&    46.25&    47.57&    105.07                   \\
1,000,000& 2350340 &   367.12&   379.53&   386.05&   394.47&   405.85 & 1027.29                  \\
		
		10,000,000 & 41957790 &  476.62 & 489.98 & 497.17 & 505.63 & 515.77 & 10229.49                \\ \bottomrule
\end{tabular}
	\centering
	\caption{Performance for L2 - in the original space - $1104$ bytes}
	\label{perftable_Original}
	\begin{tabular}{@{}r|r|rrrrr|r@{}}
		\toprule
		Datasize                      & Index Build Time & \multicolumn{5}{c|}{Query Time for knn (ms)} & Brute Force time \\
		& ms               & 1       & 2       & 3      & 5      & 10     & ms                      \\ \midrule
		100 &                      20 &     0.37&     0.15&     0.10&     0.10&     0.08&    0.84 \\
		1,000 &  80 &     0.59&     0.60&     0.60&     0.60&     0.69&     1.17                    \\
		10,000 &   1626 &     5.76&     5.14&     6.05&     6.28&     6.50&     6.09                    \\
		100,000 & 122115 &    44.63&    48.09&    49.38&    51.03&    56.51&    58.39                    \\
1,000,000& 7984466 &   415.72&   441.81&   449.46&   453.83&   470.15 & 667.02                \\
		
		10,000,000 & 97797803 &  459.51&   486.25&   496.73&   508.85&   526.57 & 19982.22                \\ \bottomrule
\end{tabular}
\end{table*}

\fi

\ifOmitAdditionalIRResult

\else

\section{Other Examples of Image Retrieval Demonstration}
\label{sec:AdditionalExperimentalResultImageRetrieval}

Here, we show examples of image retrieval demonstration 
(other than the one in 
\ifWithSectionNumber
Section~\ref{sec:Experiment.Image}).
\else
the main text).
\fi
Figures~\ref{fig:image_experiment_cream_dog} and \ref{fig:image_experiment_white_dog}
show the retrieved dog images according to the MIPS score for the dog classifier (first row),
and those according to the combined L2+MIPS score  (second row) for different query images.
Similarly,
Figures~\ref{fig:image_experiment_red_car} and \ref{fig:image_experiment_blue_car}
show the retrieved vehicle images according to  the MIPS score for a vehicle classifier,
and those according to the combined L2+MIPS score for different query images.

 \def\figsize{0.8\textwidth}

\begin{figure*}[t]
 \centering
 \includegraphics[width=\figsize]{examples/cream_dog.pdf}
 \caption{First row: Top dog images according to the MIPS score. Second row: Top dog images according to the combined L2+MIPS score ($\gamma^{(1)} = 0.8$ and $\lambda^{(2)} = 0.2$).}
 \label{fig:image_experiment_cream_dog}
%
\vspace{15mm}
%
 \centering
 \includegraphics[width=\figsize]{examples/white_dog.pdf}
 \caption{First row: Top dog images according to the MIPS score. Second row: Top dog images according to the combined L2+MIPS score ($\gamma^{(1)} = 0.8$ and $\lambda^{(2)} = 0.2$).}
 \label{fig:image_experiment_white_dog}
\end{figure*}

\begin{figure*}[t]
 \centering
 \includegraphics[width=\figsize]{examples/red_car.pdf}
 \caption{First row: Top vehicle images according to the MIPS score. Second row: Top vehicle images according to the combined L2+MIPS score ($\gamma^{(1)} = 0.8$ and $\lambda^{(2)} = 0.2$).}
 \label{fig:image_experiment_red_car}
%
\vspace{15mm}
%
 \centering
 \includegraphics[width=\figsize]{examples/blue_car.pdf}
 \caption{First row: Top vehicle images according to the MIPS score. Second row: Top vehicle images according to the combined L2+MIPS score ($\gamma^{(1)} = 0.8$ and $\lambda^{(2)} = 0.2$).}
 \label{fig:image_experiment_blue_car}
\end{figure*}

\fi

\ifReview
\else
\subsubsection*{Acknowledgments}
This work was supported by
the German Research Foundation (GRK 1589/1)
by the Federal Ministry of Education and Research (BMBF) 
under the project Berlin Big Data Center (FKZ 01IS14013A).

\fi

\ifCLASSOPTIONcaptionsoff
  \newpage
\fi



%

\bibliography{MachineLearning}
\bibliographystyle{IEEEtran}

%

\ifOmitBiography
\else

\begin{IEEEbiography}{}
is a 
\end{IEEEbiography}

\begin{IEEEbiography}[{\includegraphics[width=1in,height=1.25in,clip,keepaspectratio]{biography/klaus}}]{Klaus-Robert M\"uller}
studied physics at University of Karlsruhe,
Germany, from 1984 to 1989 and received the
Ph.D. degree in computer science from University of Karlsruhe in 1992.
He has been a Professor of computer science at
Technische Universit{\"a}t Berlin, Berlin, Germany,
since 2006. At the same time he has been the
Director of the Bernstein Focus on Neurotechnology
Berlin until 2013; from 2014 he has been Co-director of the Berlin Big Data Center. 
After completing a postdoctoral position
at GMD FIRST in Berlin, he was a Research Fellow at the University of
Tokyo from 1994 to 1995. In 1995, he founded the Intelligent Data
Analysis group at GMD-FIRST (later Fraunhofer FIRST) and directed it
until 2008. From 1999 to 2006, he was a Professor at the University of
Potsdam.
Dr. M{\"u}ller was awarded the 1999 Olympus Prize by the German
Pattern Recognition Society, DAGM, and, in 2006, he received the SEL
Alcatel Communication Award. In 2014 he received the Berliner
Wissenschaftspreis des regierenden B{\"u}germeisters. In 2012, he was
elected to be a member of the German National Academy of Sciences-
Leopoldina. His research interests are intelligent data analysis, machine
learning, signal processing, and brain computer interfaces.
\end{IEEEbiography}

\begin{IEEEbiography}[{\includegraphics[width=1in,height=1.25in,clip,keepaspectratio]{biography/nakajima}}]{Shinichi Nakajima}
is a senior researcher in Berlin Big Data Center, Machine Learning Group, Technische Universit\"at Berlin.  He received the master degree on physics in 1995 from Kobe university, and worked with Nikon Corporation until September 2014 on statistical analysis, image processing, and machine learning.  He received the doctoral degree on computer science in 2006 from Tokyo Institute of Technology.  His research interest is in theory and applications of machine learning, in particular, Bayesian learning theory, computer vision, and data mining.
\end{IEEEbiography}





\fi

\end{document}